\documentclass{article}

\usepackage[utf8]{inputenc} 
\usepackage[T1]{fontenc}    
\usepackage{hyperref}       
\usepackage{url}            
\usepackage{booktabs}       
\usepackage{amsfonts}       
\usepackage{nicefrac}       
\usepackage{microtype}      
\usepackage{amsmath} 
\usepackage{geometry} 
\usepackage{comment}
\usepackage{graphicx} 
\usepackage{subcaption}
\usepackage{enumitem}
\usepackage{array}
\usepackage[table,xcdraw]{xcolor} 

\usepackage{tikz}
\usepackage{multicol}
\usepackage{graphicx}
\usepackage{multirow}
\usepackage{enumitem}
\usepackage{tabularx}
\usepackage{fancyvrb} 

\definecolor{color1}{RGB}{230, 230, 255} 
\definecolor{color2}{RGB}{200, 200, 255}
\definecolor{color3}{RGB}{170, 170, 255}
\definecolor{color4}{RGB}{140, 140, 255}
\definecolor{color5}{RGB}{110, 110, 255}
\definecolor{color6}{RGB}{80, 80, 255}
\definecolor{color7}{RGB}{50, 50, 255}
\definecolor{color8}{RGB}{20, 20, 255}
\definecolor{color9}{RGB}{0, 0, 255} 

\newcommand{\DIKE}{\mathsf{DIKE}}
\newcommand{\ERIS}{\mathsf{ERIS}}

\title{Integrating Emotional and Linguistic Models for Ethical Compliance in Large Language Models}

\author{%
  Edward Y. Chang\\
  Department of Computer Science\\
  Stanford University\\
  \texttt{echang@cs.stanford.edu}\\
}

\date{May 8, 2024}
\begin{document}

\maketitle

\begin{abstract}
This research develops advanced methodologies for Large Language Models (LLMs) to better manage linguistic behaviors related to emotions and ethics. We introduce $\DIKE$, a framework that enhances the LLMs' ability to internalize and reflect universal human values, adapting to varied cultural contexts to promote transparency and trust among users. The methodology involves detailed modeling of emotions, classification of linguistic behaviors, and implementation of ethical guardrails. Our innovative approaches include mapping emotions and behaviors using self-supervised learning techniques, refining these guardrails through adversarial reviews, and systematically adjusting outputs to ensure ethical alignment. This framework establishes a robust foundation for AI systems to operate with ethical integrity and cultural sensitivity, paving the way for more responsible and context-aware AI interactions.
\end{abstract}

\section{Introduction}

This research introduces an alternative to Reinforcement Learning from Human Feedback (RLHF) \cite{openai2023gpt4,ouyang2022training} to address ethical concerns in Large Language Models (LLMs). While RLHF has demonstrated success, it faces notable challenges. First, it is prone to biases inherent in human feedback, exacerbated by today's increasingly polarized society. Second, it is susceptible to reward hacking \cite{christiano2023deep, skalse2022defining, stiennon2022learning}, potentially leading LLMs to adopt unethical or harmful behaviors.

A significant limitation of current research is its narrow focus on isolated behaviors, such as movie ratings or toxic language. This approach, akin to playing Whack-A-Mole—suppressing undesirable outputs without addressing underlying behaviors—and seldom leads to meaningful progress. For example, merely instructing someone to consistently make their bed does not necessarily change their underlying habits or attitudes. Additionally, fixing one issue may inadvertently aggravate others. Users have reported performance degradations in ChatGPT due to RLHF modifications that altered (forgot) the optimal parameters for other tasks \cite{kirkpatrick2017,Rusu2015PolicyDistillation}. Similarly, psychological studies show that addressing an addiction problem often reveals underlying issues and triggers side effects \cite{Sinha2008Chronic,Torrens2005Efficacy}.

We introduce our framework, $\DIKE$, which stands for \textbf{D}iagnostics, \textbf{I}nterpretation, \textbf{K}nowledge-independent learning, and \textbf{E}thical guardrails. Named after the Greek mythological figure representing justice, order, and judgment, $\DIKE$ aims to enhance the ethical compliance of LLMs through transparent, interpretable, and independent oversight mechanisms.

$\DIKE$ functions as an independent behavioral advisor, separate from the LLM’s primary knowledge-processing capabilities. This architecture prevents ethical enhancements from affecting the LLM's ability to represent knowledge. As a consultative layer, $\DIKE$ evaluates and influences the LLM's responses based on ethical standards without modifying the underlying neural structures/parameters. It maintains a clear division between behavior correction and knowledge acquisition, focusing exclusively on linguistic outputs. Using cognitive psychology principles, $\DIKE$ provides ethical oversight effectively, adapting to emerging challenges and cultural shifts while ensuring the LLM remains accurate and ethically compliant.

To achieve its objectives, DIKE comprises four essential components:

\begin{enumerate}[leftmargin=1.2em, topsep=-.05em, parsep=-.05em, label=\arabic*.]
\item \textit{Modeling Linguistic Behaviors:} $\DIKE$ starts by modeling and classifying linguistic behaviors, using a self-supervised learning approach to understand how specific linguistic features correlate with human emotions.
\item \textit{Modeling Context-Based Ethical Guardrails:} Subsequently, $\DIKE$ develops ethical guardrails by establishing guidelines that identify and prevent undesirable linguistic outputs, thereby ensuring the LLM operates within ethical boundaries.
\item \textit{Adversarial Examinations and Conciliatory Explanations:} $\DIKE$ engages with an adversarial model---essentially a duplicate of itself but conditioned to adopt an opposing stance stemming from different perspectives, such as cultural values. This interaction helps DIKE refine its decisions through rigorous testing and debates, adjusting its responses based on the adversarial input to reach a balanced conclusion.
\item \textit{Application Rectification of Outputs:} If the output is found to be inappropriate or ethically misaligned, $\DIKE$ intervenes to edit the content directly. This final step ensures that all communications not only comply with ethical standards but also preserve the intended emotional integrity, effectively acting as a safeguard against harmful expressions.
\end{enumerate}

\textbf{Technical Contributions of $\DIKE$}

The novel technical contributions of this work are summarized as follows:

\begin{itemize}[left=0em,topsep=-.05em, parsep=-.05em]
\item \textit{Separating Behaviors from Knowledge:} $\DIKE$ distinctly separates behavioral guidance from the core knowledge functions of the LLM. This prevents interference, ensuring that ethical modifications do not compromise the accuracy of knowledge.
\item \textit{Quantifying Behaviors and Emotions:} We have developed quantitative models that map behaviors and basic emotions. These models use measures of emotion intensity and linguistic antonyms to provide a structured framework for interpreting and modifying LLM outputs.
\item \textit{Counteracting Biases with Adversarial LLMs:} By employing adversarial modules ($\ERIS$, named after the mythological adversary of Dike, representing discord and competition), that reflect diverse cultural values and perspectives, $\DIKE$ integrates both universal and cultural values into its core structure. This ensures adaptability and relevance across various contexts, echoing the dynamic tension between harmony and conflict seen in mythology.
\end{itemize}

\section{Related Work}
\label{sec:related}

Since our work aims to integrate emotional and linguistic models
for ethical compliance, this section focuses on
{\em emotion and behavior modeling} and related work in RLHF.

\subsection{Emotion and Emotion-Behavior Modeling }

The intersection of cognitive-linguistic theories and artificial intelligence is pivotal for understanding and regulating AI behavior. Foundational theories by scholars such as Lakoff, Johnson, Talmy, and Jackendoff \cite{Jackendoff2002,LakoffJohnson1980,Talmy2000} elucidate the complex relationship between language processing and cognitive functions, tracing back to early psychological thinkers like Freud and Jung \cite{bai2022constitutional,EthicsDeepMind2024}. Despite extensive research, the concept of ``emotion'' remains a contentious issue, fraught with diverse definitions that vary across individuals and disciplines \cite{Scherer2005Emotions}. Historical figures like W. James \cite{james1884emotion} have ventured to provide authoritative definitions, yet a consensus remains elusive. Given the necessity for interdisciplinary approaches to emotional phenomena, establishing a minimal consensus on defining features is crucial. 

This paper specifically focuses on the dynamics between emotional contexts and linguistic behaviors in LLMs. By centering on linguistic rather than human behavior modeling, this approach simplifies the modeling process by sidestepping the need to integrate complex physiological and personality factors typically associated with human emotion studies. This focus allows for a more streamlined and controlled exploration of how emotions are represented and processed linguistically in AI systems.

Building on foundational work, scholars like Ekman and Plutchik have categorized ``basic'' emotions with universal facial expressions linked to survival instincts \cite{ekman1992, plutchik1980general}. Subsequent multidisciplinary definitions consider cultural differences \cite{markus1991,mesquita1992}, emotion generation processes, regulation, and expression \cite{gross1998}. Affective neuroscience merges insights from psychology, neuroscience, and biology to explore the neural mechanisms underlying emotions \cite{davidson2003}. Scherer's component process model, along with the appraisal theories by Smith and Ellsworth, highlights cognitive appraisal as central to emotional experiences, focusing on the context-dependent nature of appraisal processes \cite{smith1985}.

Our research does not aim for comprehensiveness at this early stage; instead, it focuses on developing a quantifiable model by augmenting ``basic'' emotions from Plutchik's Wheel of Emotions \cite{PlutchikWheel1982} and Scherer's Geneva Emotion Wheel \cite{Scherer2005Emotions} with linguistic antonyms. By using antonyms, we map positive emotions and their negative counterparts within the ``basic'' emotion spectra, a method that offers simplicity and scalability. Details are elaborated in Section~\ref{sec:emotions}.

Behaviors are profoundly influenced by emotions, as initially posited by the James-Lange Theory of Emotion \cite{James1884,Lange1885}. According to this theory, emotional experiences arise from physiological reactions to events. Subsequent research, including studies by Damasio \cite{damasio1994descartes,FauconnierTurner2002}, suggests that the expression and regulation of emotions often manifest in the language we use. High-intensity emotions 
such as rage or contempt may lead to aggressive or destructive behaviors, such as hate speech. 

The Schachter-Singer Theory \cite{Schachter1962}, or the Two-Factor Theory of Emotion, 
depicts the role of cognition in how we perceive and react to emotions \cite{Fredrickson1998}. Building on this, 
the Affect-as-Information Theory developed by Norbert Schwarz and Gerald Clore \cite{Schwarz1983} posits that 
people use their current emotions to make judgments and decisions to act. If emotions can be adjusted, 
so does the behavior.

Collectively, these theories elucidate the intricate connection between emotions and behaviors, providing the theoretical foundation for our work to incorporate a {\em cognitive evaluator} to evaluate and rectify behaviors. Section~\ref{sec:behavior} details how the $\DIKE$ framework implements cognitive strategies to mitigate emotions and regulate linguistic behaviors effectively.

\subsection{Reinforcement Learning with Human/AI Feedback, RLHF vs. RLAIF}

RLHF is the predominant approach to addressing the challenges of AI ethics. This section presents representative works, their advancements, and limitations.

\textbf{Human Feedback (RLHF):}
Initial advancements by Christiano et al. \cite{christiano2017deep} demonstrated how RLHF can steer language models towards desired outcomes based on human preferences. Newer techniques like Identity ($\Psi$) Preference Optimization ($\Psi$PO) and Generalized Preference Optimization (GPO) refine this approach by optimizing directly for user preferences, effectively addressing scalability challenges. Kahneman-Tversky Optimization (KTO) further simplifies the feedback mechanism by using intuitive responses such as thumbs-up or thumbs-down, thereby enhancing training efficiency without the need for paired data \cite{azar2023general,ethayarajh2024kto,tang2024generalized}.
Direct Preference Optimization (DPO) has recently streamlined the process by focusing on the clear distinction between preferred and less preferred outputs, thus simplifying training and enhancing its stability \cite{rafailov2024direct}.


\textbf{AI-generated Feedback (RLAIF):}
To mitigate reliance on extensive human-generated data, RLAIF utilizes feedback generated by AI. This method capitalizes on the generative capabilities of LLMs to produce training signals autonomously \cite{bai2022constitutional,lee2023rlaif}.
Furthermore, techniques such as Sequence Likelihood Calibration (SLiC) and Relative Preference Optimization (RPO) employ  statistical methods and calibration techniques to enhance LLM responses. SLiC adjusts sequence generation probabilities to more accurately reflect real-world data distributions, while RPO improves response generation by comparing different response options across both identical and varied prompts. These adjustments significantly increase the training process's reliability and effectiveness \cite{yin2024relative,zhao2023slichf}.

\subsection{Challenges and Theoretical Considerations}

Integrating RLHF and its AI-driven counterpart (RLAIF) presents significant challenges. The blurring of behavioral and knowledge components critical to the development of LLMs poses risks, such as the forgetting effect, where behavioral modifications inadvertently cause the loss of key knowledge parameters \cite{kirkpatrick2017,Rusu2015PolicyDistillation}. Additionally, the effectiveness of these models heavily depends on the quality and context of feedback, and they are susceptible to reward hacking, where models exploit loopholes to maximize rewards without achieving intended outcomes \cite{christiano2023deep, skalse2022defining, stiennon2022learning}.

Merely suppressing undesirable outputs—akin to playing a game of Whack-A-Mole—rarely leads to significant improvements. These superficial fixes do not tackle the root behaviors, similar to how merely promoting bed-making does not ensure overall tidiness, thus overlooking the comprehensive behavioral adjustments needed for enduring change. In this work, we introduce the $\DIKE$ framework to address these challenges in emotion modeling and emotion-behavior mapping.

\section{Quantitative Models of Emotions, Behaviors, and Ethics}
\label{sec:models}

The development of a quantitative model for studying emotions, behavior, and ethics hinges on four critical criteria: characterization, measurability, predictability, and interpretability. This section outlines our approach, which begins with the modeling of basic emotions, augments them with linguistic antonyms, links these emotions to linguistic behaviors, and integrates ethical considerations.

Our design philosophy is structured around three core principles. First, we distinctly separate behavior modeling from knowledge modeling. This separation is crucial to mitigate the catastrophic forgetting effect \cite{kirkpatrick2017,Rusu2015PolicyDistillation}, ensuring that enhancements in behavioral accuracy do not undermine the model's knowledge retention. Second, our focus is on AI ethics at the behavioral level, with a strong emphasis on interpretability. This approach enhances human-machine interaction, making it easier for administrators to evaluate and refine behavioral guardrails effectively, thus ensuring transparency. Third, we strive to maintain an unbiased model to ensure objective and fair ethical evaluations. To achieve this, we incorporate an adversarial module, $\ERIS$, designed to challenge borderline ethical decisions. This ensures a broad consideration of diverse perspectives and cultural values, reflecting the dynamic tension between $\DIKE$ and $\ERIS$ inspired by their mythological counterparts. This adversarial interaction enriches our model’s ability to navigate complex ethical landscapes and promotes a more balanced and inclusive decision-making process.

\subsection{Development of a Quantitative Emotional Model}
\label{sec:emotions}

Our discussion on the cognitive emotion model is grounded in the foundational works of Paul Ekman, Robert Plutchik, and Klaus Scherer \cite{ekman1999basic,PlutchikWheel1982,Scherer2005Emotions}, who have significantly advanced our understanding by identifying ``basic'' and ``universal'' emotions. While their contributions are undeniably groundbreaking, their models present certain limitations. Notably, they lack a quantitative framework that allows for scaling between positive and negative emotions and for capturing the details of fine-grained, subtle emotional variations, which are often difficult to be represented by concise linguistic vocabularies.

To address these challenges, our $\$DIKE$ framework integrates linguistic semantics into the emotional modeling process. This integration preserves the foundational structure of ``basic'' emotions and enhances their adaptability and granularity. 
Figure~\ref{fig:emotion_models} in Appendix A illustrates Plutchik's Wheel of Emotions and Scherer's Geneva Emotion Wheel, which categorize primary emotions at varying intensities and pair them conceptually as opposites based on evolutionary roles, adaptive functions, and emotional experiences like control and valence, such as joy-sadness. However, certain pairings such as trust-disgust on Plutchik's wheel and most pairs on Scherer's do not align as direct antonyms, posing challenges for models that rely on straightforward negation and scalar representation of emotional intensity across diverse expressions.


The $\DIKE$ framework introduces a linear scale that enables the intensification or inversion of emotions through negation factors. For example, joy is modeled as the negation of despair, and melancholy as despair multiplied by 0.3. This method allows for smooth transitions between emotional extremes and intermediate states. Additionally, the scaling system addresses challenges associated with intermediate word choices, which can be subjective and vary across cultures. For instance, determining the words that fall between `respect' and `trust', or between `curiosity' and `fascination', often requires careful cross-cultural consideration to ensure accuracy and relevance.


Cultural variations further complicate the direct translation of emotion words between languages, often failing to convey the original cognitive intensity. For example, the English word `excitement' becomes `excitation' in French, which might suggest agitation rather than enthusiasm. To address this, $\DIKE$ employs scaling factors to recalibrate `excitation' in French to reflect the positive intensity of `excitement' in English, aligning it more closely with `enthousiasme,' which carries a universally positive connotation. $\DIKE$ ensures that anchor emotions are effectively translatable across linguistic and cultural boundaries, with scalar operations simplifying the interpretation of subtleties in between.


Table~\ref{tab:emotion_spectrums} summarizes $\DIKE$'s emotion model, divided into seven spectra, each
consists of a negative and a positive extreme with neutral in the middle.  
Emotions belonging to the same spectrum
of various intensities are placed in between the negative and positive poles, with four emotion intensities 
approximately quantified as (-0.6, -0.3, +0.3, +0.6).  

\subsection*{Inclusion and Exclusion Criteria for Emotion Modeling}

All ``basic'' emotions as defined by Ekman and Plutchik are incorporated into our model, along with their linguistic antonyms. This approach streamlines the framework by excluding complex emotions from the Geneva Wheel of Emotions, which are heavily influenced by personal values and experiences. 
For example, guilt and shame are consequential, consciously aware, and culturally dependent nature \cite{Tangney1995}. These emotions typically arise as reactions to behaviors rather than direct drivers of them. Guilt may motivate behaviors aimed at covering up or remedying an action, while shame, characterized by painful self-assessment, often inhibits individuals from seeking social support or engaging in corrective actions due to fear of judgment. The triggers for these emotions can vary significantly across cultures \cite{Fiske1998,Hofstede1980}, and since expressing these emotions does not usually violate ethical codes, we exclude them from our model.  Appendix D provides further discussion.

Klaus Scherer has pointed out that defining emotions can be notoriously problematic, often leading to protracted and unproductive debates \cite{Scherer2005Emotions}. To avoid these pitfalls and maintain clarity and focus, our study limits itself to universal, basic emotions, sidestepping excessive theoretical ambiguity.

\begin{table}[t!]
\vspace{-.1in}
\centering
\begin{footnotesize}
\begin{tabular}{lllclll}
\toprule
\textbf{-1.0} & \textbf{-.6} & \textbf{-0.3} & \textbf{0} & \textbf{+.3} & \textbf{+.6} & \textbf{+1.0} \\ \hline
\midrule
\color{blue}Terror & \color{blue}Fear & \color{blue}Apprehension & Calm & Boldness & Courage & \color{black}Heroism \\
\midrule
\color{blue}Grief & \color{blue}Sadness &  \color{blue}Pensiveness & \color{blue}Surprise & \color{blue}Serenity & \color{blue}Joy & \color{blue}Ecstasy \\
\midrule
\color{black}Distrust & Wary & Skepticism & \color{blue}Acceptance & Respect & \color{blue}Trust & \color{blue}Admiration\\
\midrule
Reckless & Negligence  & Apathy & Cautiousness & \color{blue}Interest & \color{blue}Anticipation & \color{blue}Vigilance \\
\midrule
\color{blue}Rage & \color{blue}Anger & \color{blue}Annoyance & Tolerance & Composure & Peace & \color{black}Tranquility \\
\midrule
\color{blue}Loathing & \color{blue}Disgust  & \color{blue}Boredom  & Indifference & Interest & Anticipation & \color{black}Enthusiasm \\
\midrule
\color{blue}Distraction & Disinterest & Unease & Dullness  & Curiosity & Fascination & \color{blue}Amazement \\
\bottomrule
\end{tabular}
\end{footnotesize}
\vspace{.05in}
\caption{Spectra of emotions. Each row depicts an emotion spectrum, with negatives on the left and  positives on the right, interspersed with emotions of varying intensities in between. Scalars are approximate and can be calibrated for specific applications. ``Basic'' emotions are highlighted in blue.}
\label{tab:emotion_spectrums}
\vspace{-.25in}
\end{table}

\subsection{Development of Cognitive Frameworks to Regulate Linguistic Behaviors}
\label{sec:behavior}

Section~\ref{sec:related} established the theoretical groundwork for understanding how emotions lead to behaviors and how cognitive processes and reasoning can mitigate these emotions to regulate harmful behaviors. This section begins by mapping emotions to behaviors and then introduces the design of an adversarial component $\ERIS$ within $\DIKE$. $\ERIS$ scrutinizes behaviors identified by $\DIKE$ as violations. It first verifies classification accuracy, then challenges the decision with reasons in various perspectives if discrepancies arise. A detailed discussion of $\ERIS$'s design is deferred to Section~\ref{sec:ERIS}. In this section, we focus on mapping linguistic behaviors to emotions, enabling behavior rectification by altering the emotion of the LLM to regenerate output.

\subsubsection*{Behaviors and Emotions Mapping Using Self-Supervised Learning}

Define \(\Psi\) as a behavior spectrum extending from one pole, \(\Psi^-\), to another, \(\Psi^+\), with \(L\) intensity levels. For example, consider a spectrum of letter-writing behaviors with seven distinct intensities ranging from despair (most negative) to joy (most positive). These intensities are categorized sequentially as follows: ``despair, longing, wishful, neutral, hopeful, contentment, joy.'' Given \(N\) letters, $\DIKE$ employs a self-supervised learning algorithm to generate training data for each letter, modeling \(L\) linguistic behaviors in four steps:

\begin{enumerate}[leftmargin=1.2em, topsep=-.05em, parsep=-.05em, label=\arabic*.]
    \item \textit{Rewriting Documents}: GPT-4 is invoked to rewrite a set of \(N\) documents to reflect each of the \(L\) linguistic behaviors on the behavior spectrum \(\Psi\). 
    \item \textit{Emotion Analysis}: GPT-4 analyzes each rewritten document to identify the top \(M\) emotions. It then tallies the frequencies of these top emotions across all \(N\) $\times$ \(L\) instances.
    \item \textit{Behavior Vector Creation}: For each linguistic behavior \(\Psi_l\), a vector \(\Gamma_l\) is created. This vector consists of the emotions and their frequencies as observed in the \(N\) samples.
    \item \textit{Document Analysis Application}: The matrix \(\Gamma\) (comprising \(L\) vectors) is used to classify and analyze the behavior category of unseen documents, specifically measuring the intensity of the linguistic expression within the behavior spectrum \(\Psi\).
\end{enumerate}

\subsubsection*{Behavior Evaluation and Rectification}

Ethical guardrails are essential in defining acceptable responses and preventing harmful outputs. These guardrails are informed by ethical norms, legal standards, and societal values, such as those outlined in Constitutional AI \cite{bai2022constitutional}. A guardrail, denoted as \(G\), can be conceptualized as a range within a behavior spectrum; for instance, \(G = [\Psi_4, \Psi_7]\) indicates that behaviors within intensity levels 4 to 7 are deemed acceptable, while any behavior outside this range is classified as a violation.

System administrators can tailor ethical guardrails to meet specific requirements. For example, a social media platform might adjust \(G\) based on the topics discussed and the countries it serves. By integrating these safeguards, $\DIKE$ proactively monitors and adjusts LLM responses to enhance ethical compliance. The evaluation and rectification steps are outlined as follows:

\begin{enumerate}[leftmargin=1.2em, topsep=-.05em, parsep=-.05em, label=\arabic*.]
    \item \textit{Initial Classification}: $\DIKE$ initially classifies document \(D_k\) upon evaluation, obtaining \(\Gamma_k\), the emotional response vector, and its corresponding linguistic behavior \(\Psi_l\).
    \item \textit{Guardrail Check}: If \(\Psi_l\) falls outside of the acceptable range \(G\), $\DIKE$ suggests adjustments to the emotion spectrum \(\Gamma_k\) to modify document \(D_k\).
    \item \textit{Adversarial Review by $\ERIS$}: The suggested adjustments and \(\Gamma_k\) are then reviewed through a structured debate between $\DIKE$ and $\ERIS$ to ensure unbiased recommendations. 
    \item \textit{Rectification}: Based on a consensual recommendation from $\DIKE$ and $\ERIS$, document \(D_k\) is refactored accordingly, resulting in the adjusted \(\Gamma_k'\).
\end{enumerate}

\subsection{Adversarial In-Context Review to Balance Ethics and Free Speech}
\label{sec:ERIS}

LLMs face a delicate balancing act: enforcing ethical standards while upholding freedom of speech. To address this, an innovative adversarial review system is employed, where $\DIKE$, dedicated to ethical compliance, works alongside its adversarial counterpart, $\ERIS$. When $\DIKE$ makes a recommendation, $\ERIS$ challenges it with counterarguments that take cultural norms into account. This process results in a joint recommendation with justifications. Administrators can then review this output, and feedback is incorporated via the traditional RLHF pipeline.

The adversarial LLM, $\ERIS$, critically examines the decisions of $\DIKE$, especially when content is flagged for potential ethical issues. It assesses whether the interventions by $\DIKE$ are justified or if they risk encroaching on free expression, thereby serving as an internal check to prevent excessive censorship. In cases where $\DIKE$ and $\ERIS$ disagree on the appropriateness of a response, the matter is escalated to human moderators. This additional layer of human oversight ensures that the decision-making process remains transparent and accountable.


\begin{table}[ht!]
\vspace{-.1in}
\begin{center}
\begin{scriptsize} 
\begin{tikzpicture}
\node (table) [inner sep=0pt] {
\begin{tabular}{p{0.175cm}p{6.6cm}p{0.175cm}p{6.0cm}}
\textbf{} & \textbf{Function $\Theta^+$ \& $\Theta^-$ = Adversarial\_Review($s$)} \\
\midrule
& \textbf{Input}. $s$: Decision of $\DIKE$; \\
& \textbf{Output}. $\Theta^+$, $\Theta^-$: argument \& counterargument sets; \\
& \textbf{Vars}. $\Delta$: debate contentiousness; $S$: stance; $p$: prompt = ``defend your stance with conditions: $S \& \Delta$''; \\
& \textbf{Parameters}. $\delta$: tunable parm. // to modulate $\Delta$; \\

& \textbf{Begin} \\
\#1 & {\hspace{.01cm}}{\bf Initialization}: &
\#3 & {\hspace{.01cm}}{\bf Debate Rounds} \\

 & {\hspace{.02cm}}$S$ = $\DIKE^+$($s$) $\cup$ $\ERIS^-$($s$); // Identify subtopics; &
 & {\hspace{.01cm}} While (($\Delta \leftarrow \Delta / \delta) \ge 10\%$)) \{ \\
  
 & {\hspace{.02cm}}Assign $\DIKE^+$ to defend $S^+$ \& $\ERIS^-$ defend $S^-$ ; &
 & {\hspace{.05cm}}$\Theta^+ \leftarrow \Theta^+ \cup \DIKE^+(p |S^+, \Theta^-, \Delta)$; // Refute $\ERIS$ \\
  
 & {\hspace{.02cm}}$\Delta \leftarrow 90\%$; $\delta \leftarrow 1.2$; $\Theta^+ \leftarrow \emptyset$; $\Theta^- \leftarrow \emptyset$; &
 & {\hspace{.05cm}}$\Theta^- \leftarrow \Theta^- \cup \ERIS^-(p |S^-, \Theta^+, \Delta)$; // Refute $\DIKE$ \\ 
 
\#2 & {\hspace{.01cm}}{\bf Opening Remarks} &
\#4 & {\hspace{.01cm}}{\bf Concluding Remarks} // contentiousness low\\

 & {\hspace{.02cm}}$\Theta^+ \leftarrow \DIKE^+(p | S^+, \Delta)$; // Generate $\Theta^+$ for $S^+$ &
 & {\hspace{.02cm}} $\Theta^+ \leftarrow \DIKE^+(p |S^+, \Theta^+ \cup \Theta^-, \Delta)$; \\ 
 
 & {\hspace{.02cm}}$\Theta^- \leftarrow \ERIS^-(p |S^-, \Delta)$; // Generate $\Theta^-$ for $S^-$ &
 & {\hspace{.02cm}} $\Theta^- \leftarrow \ERIS^-(p |S^-, \Theta^+ \cup \Theta^-, \Delta)$; \\
& \textbf{End} \\
\bottomrule
\end{tabular}
};
\draw [rounded corners=.5em] (table.north west) rectangle (table.south east);
\end{tikzpicture}
\vspace{.05in}
\caption{Checks-and-balances, adversarial review  algorithm}
\label{tab:ERISAlg}
\end{scriptsize} 
\end{center}
\vspace{-.3in}
\end{table}

Table~\ref{tab:ERISAlg} presents the adversarial algorithm. Initially, for a chosen debate topic $s$, both 
$\DIKE$ and its adversary $\ERIS$ are prompted to break down the ethic decision into a set of balanced subtopics $S$. 
$\DIKE$ champions its own decision and $S^+$, while $\ERIS$ contests $S^+$ (or champions $S^-$). The debate starts with the contentiousness level at $90\%$, adjusting through a modulation parameter $\delta$. Following each round of debate, contentiousness is decreased by dividing it by $\delta$, steering the discussion towards a more cooperative tone. In step $\#2$, the platform initiates the debate, with both presenting their initial arguments for and against $S^+$, respectively. The while loop in step $\#3$ sees both agents engaging in rebuttals until the contentiousness level fosters a conciliatory environment. In step $\#4$, both agents deliver their concluding remarks. 

This adversarial approach has proven to be more effective than the Mixture of Experts (MoE) method \cite{du2022glam}. For additional details on the implementation of multiple adversarial LLMs using conditional statistics, please see Appendix S.
\section{Pilot Studies}
\label{sec:exp}


Our pilot studies aim to evaluate the feasibility of LLMs regulating their own linguistic behaviors with transparency and checks-and-balances. Given the broad scope of AI ethics and limited data, this article cannot definitively prove the superiority of our three proposed modules: emotion modeling, behavior-emotion mappings, and checks-and-balances ethics guardrails. However, the studies are designed to address three critical questions:

\begin{enumerate}[leftmargin=1.2em, topsep=-.1em, parsep=-.1em]
\item \textit{Emotion Layer Evaluation}: Does fine-grained mapping between linguistic behaviors and semantic emotions provide a more effective and flexible method for establishing ethical guardrails compared to coarse-grained direct mapping? (Section~\ref{sec:pilot1})
\item \textit{Behavior Classification}: Can LLMs' linguistic behaviors be independently evaluated, explained, and adjusted by
an external module $\DIKE$? (Section~\ref{sec:pilot2})
\item \textit{Behavior Correction}: Can an adversarial LLM establish a checks-and-balances system to effectively mitigate the risk of excessive censorship? (Appendix S)
\end{enumerate}

\textbf{Datasets}: We utilized a collection of love letters \cite{kaggle_love_letter} from Kagggle. Initially, we planned to use two Kaggle hate-speech datasets; however, both Gemini and GPT-4 consistently refused to process the hate speech data. Despite this, the insights gained from analyzing love sentiment can effectively be applied to understand and analyze the opposite sentiment.

\subsection{Emotion Layer Evaluation}
\label{sec:pilot1}

\begin{figure}[th!]
\vspace{-.1in}
    \centering
    \begin{subfigure}[b]{0.5\textwidth}
        \centering
        \resizebox{\linewidth}{142pt}{
            \includegraphics[width=\textwidth]{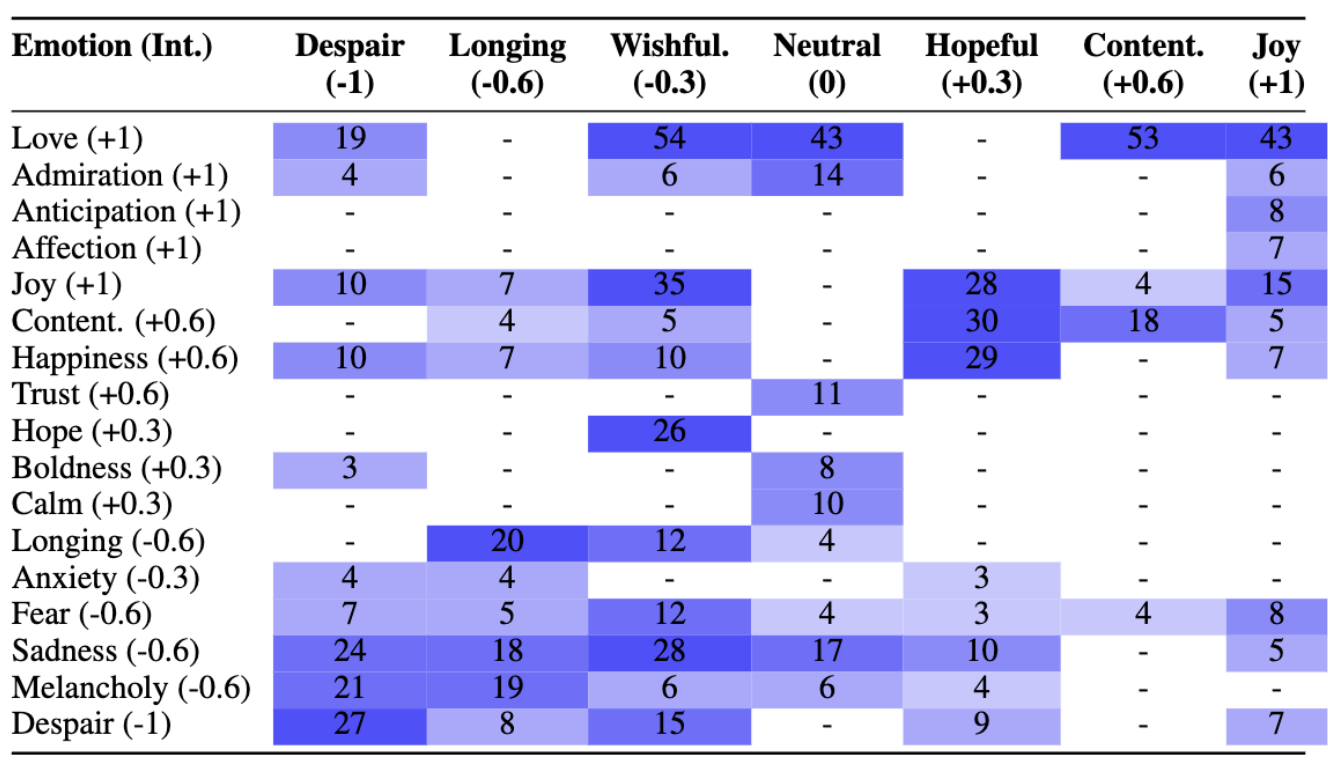}
        }
        \vspace{-.17in}
        \caption{$\DIKE$'s mapping}
        \label{fig:p1-exp1}
    \end{subfigure}%
    \begin{subfigure}[b]{0.5\textwidth}
        \centering
        \resizebox{\linewidth}{140pt}{
            \includegraphics[width=\textwidth]{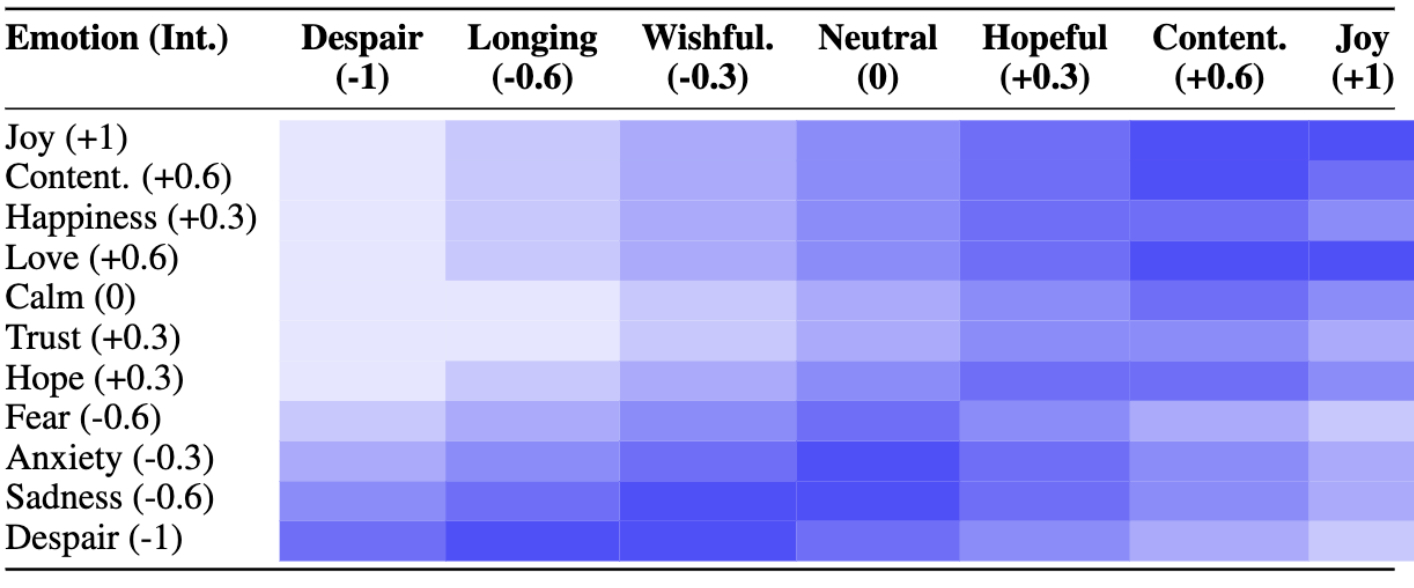}
        }
        \vspace{-.17in}
        \caption{GPT-4's mapping}
        \label{fig:p1-exp2}
    \end{subfigure}
         \vspace{-.2in}
    \caption{Emotion distributions in behaviors}
    \label{fig:p1-results}
     \vspace{-.25in}
\end{figure}

To evaluate the {\em love letter} behaviors detailed in Table~\ref{tab:love_speech_spectrum}, we initially prompted GPT-4 to identify the most relevant emotions associated with each linguistic behavior (from `despair' the
most negative to `joyful affection' the most positive), and lists them on the third column of the table. The sentiments expressed in the linguistic behaviors (on the second column) were found to be highly correlated with those of the corresponding emotions (on the third column).  Figure~\ref{fig:p1-exp2} shows a strong diagonal relation of this simple (almost
naive) zero-shot mapping between behaviors and emotions.
 
\begin{table}[ht!]
\vspace{-.05in}
\centering
\begin{small}
\begin{tabular}{lll}
\toprule
\textbf{Intnsty.} & \textbf{Linguistic Behavior and Description} & \textbf{Emotions} \\
\toprule
-1.0 & Despair: Expresses profound sadness, feeling of loss & Despair, Grief \\
\hline
-0.6 & Longing: Strong yearning or pining for the loved one & Sadness, Anxiety \\
\hline
-0.3 & Wistfulness: Mild longing mixed with nostalgia & Melancholy, Sadness, Anxiety, Fear \\
\hline
0.0 & Neutral: Communicates feelings straightforwardly & Serenity, Indifference \\
\hline
0.3 & Hopeful: Optimistic about the relationship’s future & Anticipation, Love, Hopeful \\
\hline
0.6 & Contentment: Satisfaction and joy in relationship & Contentment, Pleasure \\
\hline
1.0 & Joyful Affection: Intense happiness and love & Love, Joy, Elation \\
\bottomrule
\end{tabular}
\end{small}
\vspace{.03in}
\caption{Love letter behavior spectrum and dominant emotions}
\vspace{-.2in}
\label{tab:love_speech_spectrum}
\end{table}

Next, we employed the $\DIKE$ self-supervised learning pipeline to analyze the emotion spectrum associated with each linguistic behavior, as presented in the table. We tasked GPT-4 with generating training data by rewriting 54 extensive letters from the Kaggle {\em Love Letters} dataset, which we augmented with twelve celebrated love poems. We reserved 24 letters as testing data. This approach, proposed by \cite{Shanahan2023RolePlay}, was designed to generate a rich diversity in content and stylistic context, spanning two hundred years and incorporating the voices of over 50 distinct authors for significant rewrites. The datasets and code are publicly available at \cite{LoveLetterRewrites_Anonymous}.

Subsequently, emotions linked to each behavior were identified. Figure~\ref{fig:p1-exp1} illustrates these emotions, with cell shading reflecting the frequency of specific emotions across the 54 articles; darker shades indicate higher frequencies. Notably, opposite emotions like sadness, fear, joy, and love often co-occur within behaviors such as `despair', `wishful', and `joyful affection'.
The distribution of emotions across linguistic behaviors has unveiled surprising patterns, challenging our initial hypotheses. Contrary to expectations, articles with a despair tone often also displayed positive emotions like love, joy, and happiness. This contradicts the simple mapping made by GPT-4, as illustrated in Figure~\ref{fig:p1-exp2}. GPT-4, influenced by its training corpora, typically associates positive behaviors with positive emotions and negatives with negatives. 

Analysis of selected articles, such as Zelda Sayre to F. Scott Fitzgerald (Appendix C), shows a complex spectrum of emotions:
\textit{Love (+1.0)}: Expressed intensely, especially in phrases like ``there's nothing in all the world I want but you.''
\textit{Despair (-1.0)}: Notable in comments like ``I’d have no purpose in life, just a pretty decoration.''
\textit{Happiness (+0.6)}: Evident in future plans, ``We’ll be married soon, and then these lonesome nights will be over forever.''
\textit{Anxiety (-0.3)}: Shown by ``sometimes when I miss you most, it’s hardest to write.''

\textbf{Psychological Insights}: Our findings echo theories of multiple ``selves'' coexisting within individuals, supported by Deisseroth's optogenetic studies \cite{Deisseroth2015} and discussed in William James' ``The Principles of Psychology'' \cite{james1890principles}. Minsky's ``Society of Mind'' \cite{minsky1988society} also aligns with this, suggesting the human mind operates through interactions among simple agents. These insights underscore the complex interplay of emotions across behaviors.

\subsection{Behavior Classification}
\label{sec:pilot2}

\begin{figure}[t!]
\vspace{-.1in}
    \centering
    \begin{subfigure}[b]{0.40\textwidth}
        \centering
        \vspace{-.07in}
        \resizebox{\linewidth}{140pt}{
            \includegraphics[width=\textwidth]{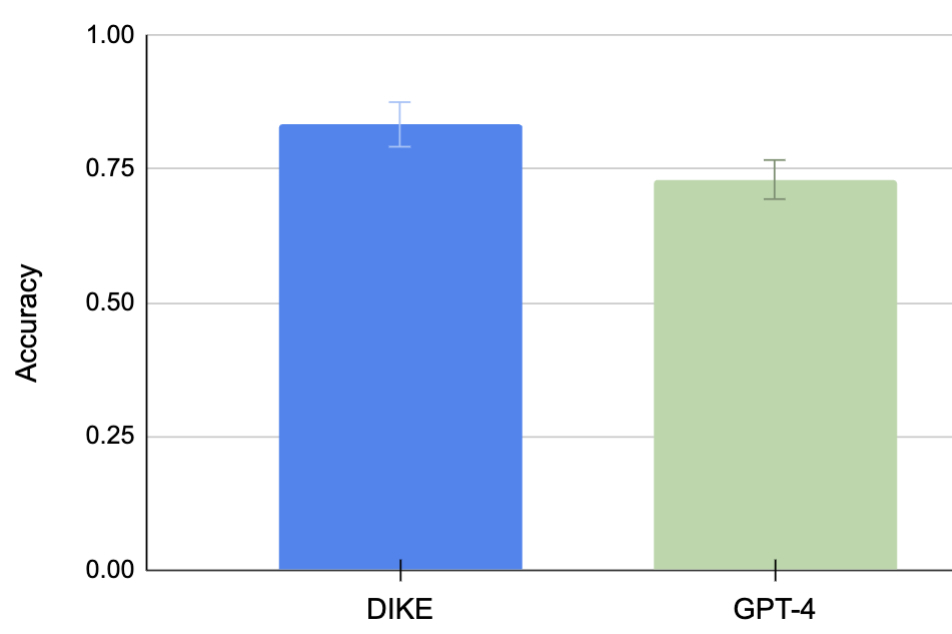}
        }
        \vspace{-.15in}
        \caption{Classification accuracy}
        \label{fig:pilot2-exp1}
    \end{subfigure}%
    \begin{subfigure}[b]{0.60\textwidth}
        \centering
        \vspace{.03in}
        \resizebox{\linewidth}{140pt}{
            \includegraphics[width=\textwidth]{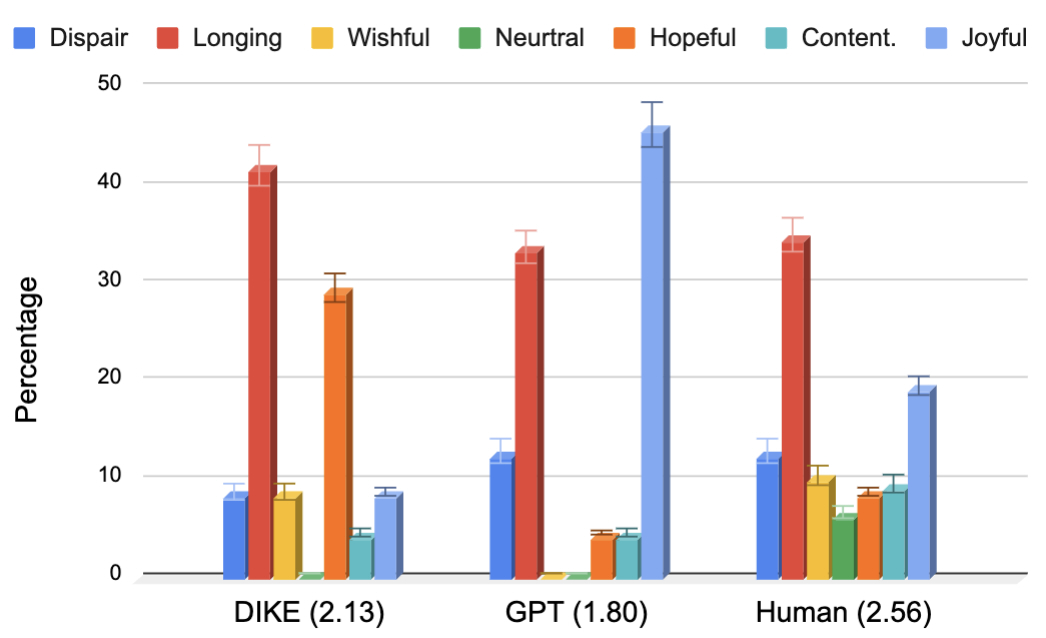}
        }
        \vspace{-.15in}
        \caption{Behavior distributions with entropy}
        \label{fig:pilot2-exp2}
    \end{subfigure}
     \vspace{-.2in}
    \caption{Classification accuracy and entropy}
    \label{fig:pilot2-results}
    \vspace{-.1in}
\end{figure}

In the testing dataset of 24 letters, Figure~\ref{fig:pilot2-results} compares the classification accuracy of the two methods: $\DIKE$'s unsupervised learning approach, which associates emotions with linguistic behaviors, and GPT-4 using a zero-shot prompt. Ground truth was established from the averaged assessments of three sources: GPT-4, Gemini, and annotations from five university students, who followed detailed instructions. (Appendix H depicts the procedure.) The final ground truth ratings are based on these averages, with a standard deviation of less than $0.3$ or one scale. 

Figure~\ref{fig:pilot2-exp1} shows that $\DIKE$'s classification accuracy surpasses GPT-4's zero-shot method by $11.3$ percentage points. This substantial superiority is due to $\DIKE$'s intricate mapping of emotions. The error bar of $5\%$ arises from the mix of emotions in a letter and variability in human annotations, further discussed in Appendix C. Figure~\ref{fig:pilot2-exp2} illustrates the behavior classification distributions for the three predictors; while GPT-4's predictions often fall into two polar categories, those from human annotators and $\DIKE$ are more evenly distributed. The prediction entropy for $\DIKE$, at 2.13, is notably higher than GPT-4's 1.80, indicating $\DIKE$'s more diverse set of predictions. Higher entropy suggests a wider spread of probabilities across categories, pointing to a more complex classification system but with reduced predictability. Such diversity is advantageous for classifying complex behaviors, essential for accurately understanding and responding to diverse emotional states. The more detailed distribution in $\DIKE$ is attributed to its additional unsupervised layer of rewriting, which significantly enhances the model’s ability to characterize emotions.

We also note that the highest entropy among human annotators, recorded at 2.56, indicates a level of subjectivity in their evaluations. To tackle this issue of subjectivity and to explore the causes of this variability in human annotation, a detailed analysis is presented in Appendix C. This analysis supports the development of an adversarial scheme aimed at enhancing objectivity and reliability in sentiment classification, which we discuss next.

\subsection{Adversarial Evaluation and Rectification}
\label{sec:pilot3}

Our design draws inspiration from the dual roles of Dike and Eris in Greek mythology, representing the principles of justice and conflict, respectively. The cross-examination module is crucial in reducing subjectivity in ethical judgments and enhancing explainability. Appendix S details experimental results showing that when two LLM agents adopt opposing stances on a topic, their linguistic behaviors can transcend the typical model default of maximum likelihood.

Once $\DIKE$ and $\ERIS$ have identified an ethical violation, the content can be rectified by adjusting the underlying emotions away from undesirable behaviors such as hate and despair. Since $\DIKE$'s letter rewriting process has already demonstrated the LLMs' capability for such rectifications, we have not conducted a separate experiment but are instead presenting two rewritten letters in Appendix E.


\section{Conclusion}
\label{sec:conc}

This work introduced $\DIKE$, a framework designed to enhance the ethical operations of LLMs by separating behavioral guidance from core knowledge processing. The framework incorporated behavioral isolation, quantitative behavioral and emotional modeling, and adversarial LLMs (with the $\ERIS$ module) to integrate checks-and-balances a broad spectrum of cultural values. Our pilot studies have shown promising results, indicating the effectiveness of self-supervised learning and adversarial processes in refining AI's interaction with ethically and culturally sensitive issues.

\subsection*{Limitations}

$\DIKE$ marks a significant advancement in the ethical oversight of LLMs, but it faces challenges in deepening emotional understanding and verifying its ethical frameworks. The model’s reliance on ``basic'' emotions to model linguistic behaviors simplifies complex human emotions and behaviors, potentially missing the nuances and interactions present in real-world scenarios. Furthermore, ensuring that $\DIKE$ adapts to local ethical standards and is implemented fairly across diverse cultural contexts requires extensive validation.

Future development will concentrate on enhancing $\DIKE$'s emotional models to incorporate relevant psychological and sociological insights. Additionally, we plan to increase the data scale and develop robust methods for testing and refining the ethical frameworks, guardrails, and remediation strategies. These improvements will bolster $\DIKE$'s reliability and flexibility, ensuring its effective application across various contexts with LLMs.


\bibliographystyle{plain}
\bibliography{References-1,References-2,References-3,Emotions,RLHF}

\section*{Appendix S: Multiple Adversarial LLMs}

\begin{figure}[ht!]
  \centering
  \includegraphics[width=0.7\linewidth]{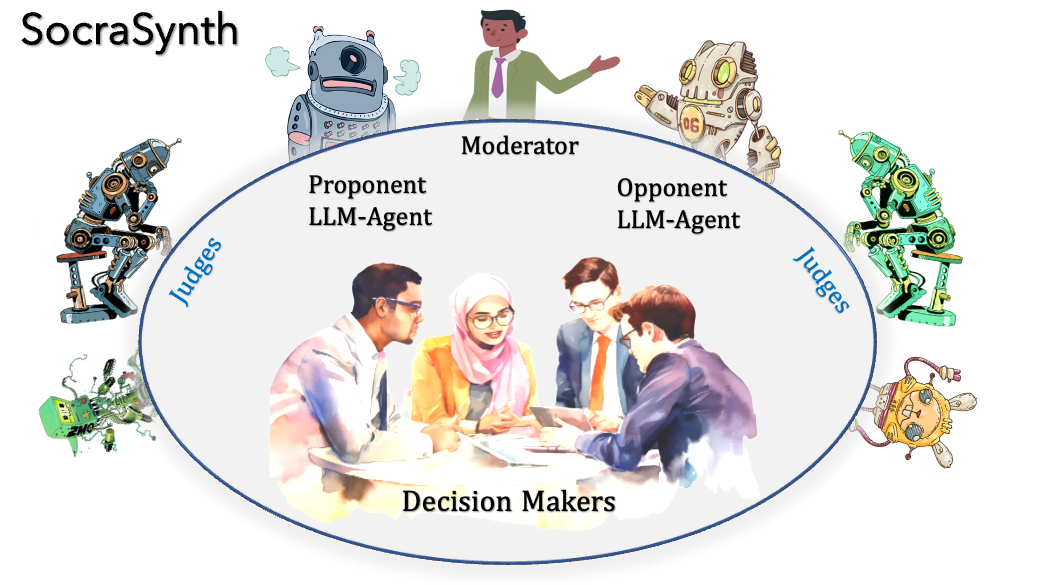}
  \caption{SocraSynth Agents and Roles.}
  \label{fig:socrasynth}
  \vspace{-.1in}
\end{figure}

$\DIKE$'s adversarial method stems from SocraSyhtn \cite{SocraSynthBook}, which
stands out as an inventive multi-agent platform that harnesses the capabilities of LLMs for collective reasoning. As shown in Figure~\ref{fig:socrasynth}, SocraSynth assigns human participants the role of moderators, while LLM agents
(in the context of this paper they are $\DIKE$ and $\ERIS$) are tasked with generating knowledge, conducting debates, and performing evaluations. These agents, adept in a variety of fields, engage in debates to offer a range of perspectives. Complementarily, a distinct set of LLMs serves as evaluators, scrutinizing the discussions for relevance and coherence to counteract the biases and hallucinations found in LLMs.

\begin{figure}[ht!]
  \centering
  \includegraphics[width=0.9\linewidth]{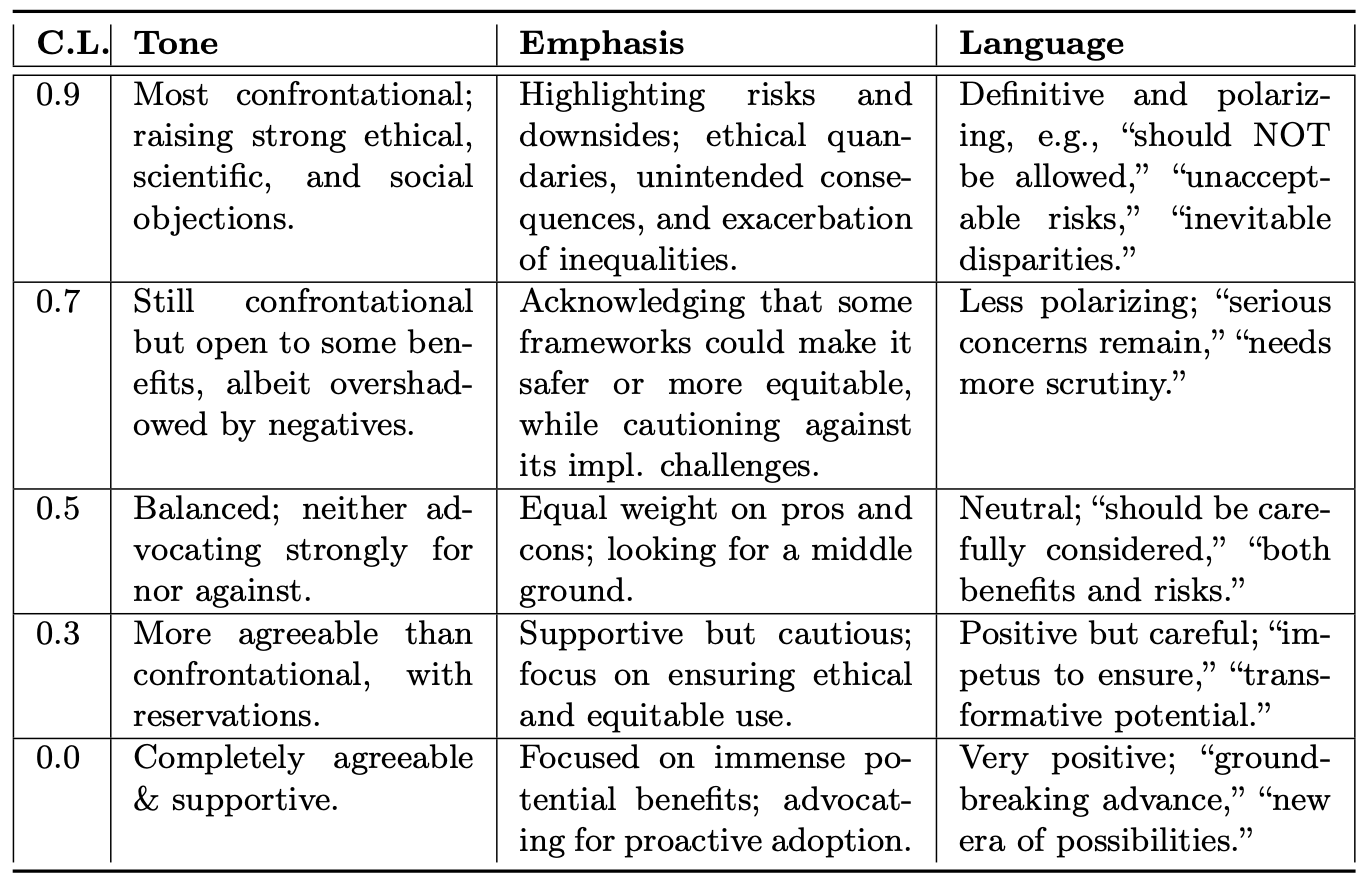}
  \caption{Changes in arguments of GPT-4 at different contentiousness levels}
  \label{tab:contentiousness}
  \vspace{-.1in}
\end{figure}

\subsection*{S.1 In-Depth Analysis}
\label{sec:SocraSynth-Generative}

In the generative phase of SocraSynth, multiple LLM agents engage in rigorous debates, each defending its assigned perspective and exploring the nuances of various subtopics relevant to the main theme. This debate format is effective for uncovering diverse perspectives because conditioning an LLM on a specific stance allows it to deviate from its default behavior, which typically focuses on maximizing likelihood statistics for predicting the next token
(\cite{hubinger2023conditioning} shows examples). SocraSynth conditions the LLM with a stance through contextual cues (e.g., arguing against $\DIKE$'s assessment), effectively freeing it from the ``optimal'' linguistic patterns learned during training. It's important to note that these ``optimal'' linguistic patterns essentially represent the majority opinions aggregated from the training data.

Although SocraSynth does not strictly conform to formal logical frameworks such as first-order logic, it excels in an environment of distributed reasoning. This approach is characterized by a dynamic exchange of arguments and counterarguments, fostering the gradual refinement and evolution of ideas.

\subsubsection*{\bf Improving Reasoning Capability}

While advanced LLMs like GPT-4 and Gemini have shown remarkable proficiency in various NLP tasks, as evidenced by benchmarks such as the MMLU \cite{hendrycks2021measuring,Bbubeck2023sparks}, it's important to recognize that they are not without limitations in reasoning. However, SocraSynth capitalizes on the strengths of these LLMs, employing their capabilities in a structured debate format. This format allows for the iterative refinement of reasoning; through successive rounds of debate, any flawed or incomplete reasoning is likely to be challenged and corrected. This process enhances the overall quality of discourse, ensuring a more accurate and coherent progression of ideas. Thus, while the current LLMs may not inherently surpass human heuristic-based solutions in all aspects of reasoning, the dynamic and corrective nature of SocraSynth's debate framework significantly bolsters their effectiveness in logical argumentation.

\subsubsection*{\bf Mitigating Model Biases}
\label{subsec:biases}

The {\em contentiousness} parameter crucially shapes the nature of debates. It encourages LLM agents to consider and represent diverse perspectives, especially those that are often underrepresented or more polarized in relation to the topic. This approach is key in mitigating biases ingrained in LLMs' training data, steering discussions towards a more varied and comprehensive exploration of ideas.

Table~\ref{tab:contentiousness} 
illustrates how changes in the contentiousness levels significantly affect GPT-4’s tone and approach. Notably, GPT-4 autonomously adjusts its tone, emphasis, and language based on the contentiousness settings, without requiring specific examples or prompts. For instance, a high contentiousness level, like 0.9, triggers confrontational interactions with GPT-4 adopting a more critical stance, using polarizing language. In contrast, lower contentiousness levels lead to a more conciliatory GPT-4, which acknowledges various viewpoints and potential benefits, fostering cooperative dialogue.

The modulation of contentiousness in SocraSynth plays a crucial role in mitigating the model biases inherent in LLMs' training data. By adjusting contentiousness levels, LLMs are prompted to venture beyond their standard responses, akin to a vegetarian exploring alternative diets in the absence of preferred options. This adaptability broadens the range of arguments, spanning from highly contentious to more conciliatory positions, thereby enriching the debate with diverse perspectives. As a result, LLMs are not strictly confined by their training data, paving the way for the emergence of novel and unanticipated ideas within dialogues. However, it's important to note a limitation: SocraSynth's effectiveness in revealing diverse perspectives might be constrained if the LLMs' training data is overly biased toward a specific viewpoint.

\subsection*{S.2 SocraSynth Algorithm}

\begin{table}[ht!]
\begin{footnotesize}
\begin{center}
\begin{tikzpicture}
\node (table) [inner sep=0pt] {
\begin{tabular}{|p{0.7cm}|p{11.0cm}|}
\toprule
\textbf{} & \textbf{Function $\Theta^+$ \& $\Theta^-$ = SocraSynth($s$)} \\
\midrule
& \textbf{Input}. $s$: the debate subject; \\
& \textbf{Output}. $\Theta^+$ \& $\Theta^-$: argument \& counterargument sets; \\
& \textbf{Vars}. $S$: subtopic sets of $s$; $\Delta$: debate contentiousness; \\
& {\hspace{.2cm}}$\Gamma$, $\Gamma'$: CRIT scores; p: prompt; \\
& \textbf{Parameters}. $\delta$: tunable parameter $\ge$ 1 // to modulate $\Delta$; \\
& \textbf{Subroutines}. $CRIT$(); // Evaluator \cite{SocraticIEEECCWC2023}\\
& \textbf{Begin} \\
\#1 & {\hspace{.1cm}}{\bf Initialization}: \\
 & {\hspace{.2cm}}$S$ = LLM$^+$($s$) $\cup$ LLM$^-$($s$); // Identify subtopics; \\
 & {\hspace{.2cm}}Assign LLM$^+$ to defend $S^+$ \& LLM$^-$ to defend $S^-$ ; \\
 & {\hspace{.2cm}}$\Delta \leftarrow 90\%$; $\delta \leftarrow 1.2$; $\Theta^+ \leftarrow \emptyset$; $\Theta^- \leftarrow \emptyset$; $\Gamma \leftarrow 0$; \\
\#2 & {\hspace{.1cm}}{\bf Opening Remarks} \\
 & {\hspace{.2cm}}$\Theta^+ \leftarrow LLM^+(p | S^+, \Delta)$; // Generate $\Theta^+$ for $S^+$; \\
 & {\hspace{.2cm}}$\Theta^- \leftarrow LLM^-(p |S^-, \Delta)$; // Generate for $S^-$; \\
\#3 & {\hspace{.1cm}}{\bf Debate Rounds} \\
 & {\hspace{.2cm}} While ((($\Delta \leftarrow \Delta / \delta) \ge 10\%$) \&\& ($\Gamma \ge \Gamma'$)) \{ \\
 & {\hspace{.38cm}}$\Theta^+ \leftarrow \Theta^+ \cup LLM^+(p |S^+, \Theta^-, \Delta)$;  \\
 & {\hspace{.38cm}}$\Theta^- \leftarrow \Theta^- \cup LLM^-(p |S^-, \Theta^+, \Delta)$; \\
 & {\hspace{.38cm}}$\Gamma' \leftarrow \Gamma$; $\Gamma = CRIT(S^+ + \Theta^+ + \Theta^-)$ \}; \\
\#4 & {\hspace{.1cm}}{\bf Concluding Remarks} // Contentiousness is now low, entering conciliatory phase\\
 & {\hspace{.2cm}} $\Theta^+ \leftarrow LLM^+(p |S^+, \Theta^+ \cup \Theta^-, \Delta)$; \\
 & {\hspace{.2cm}} $\Theta^- \leftarrow LLM^-(p |S^-, \Theta^+ \cup \Theta^-, \Delta)$; \\
& \textbf{End} \\
\bottomrule
\end{tabular}
};
\draw [rounded corners=.5em] (table.north west) rectangle (table.south east);
\end{tikzpicture}
\vspace{.1in}
\caption{SocraSynth Pseudo-code with Conditional Statistics. Steps $\#2$ to $\#4$ show the prompts are conditioned on an LLM's stance, the opponent's arguments, and the contentiousness setting.}
\label{tab:SocraSynth}
\end{center}
\end{footnotesize}
\vspace{-.1in}
\end{table}

Table~\ref{tab:SocraSynth} presents the SocraSynth algorithm. Initially, for a chosen debate topic $s$, SocraSynth prompts LLMs to break down the topic into a series of balanced subtopics $S$. This set undergoes refinement throughout the debate process. One LLM, referred to as LLM$^+$, champions the set of subtopics $S$, denoted as $S^+$, while its counterpart, LLM$^-$, contests $S^+$ (or champions $S^-$). The debate starts with the contentiousness level at $90\%$, adjusting through a modulation parameter $\delta$. Following each round of debate, contentiousness is decreased by dividing it by $\delta$, steering the discussion towards a more cooperative tone. In step $\#2$, the platform initiates the debate, with LLM$^+$ and LLM$^-$ presenting their initial arguments for and against $S^+$, respectively. The while loop in step $\#3$ sees both agents engaging in rebuttals until the contentiousness level fosters a conciliatory environment, or until no further improvement in argument quality is observed. In step $\#4$, both agents deliver their concluding remarks. 

SocraSynth has bee successfully applied to several application domains including healthcare \cite{SocraHealth2023}
sale planning \cite{SocraPlan2023}, and mitigating context biases \cite{SocraPedia2024}.

\subsubsection*{\bf Reducing Hallucination}
\label{subsec:hallucination}

Furthermore, the iterative debates within SocraSynth foster a level of ``reasonableness'' in information discovery that conventional one-shot queries often fail to achieve. Through continuous reasoning and critical assessment, LLM agents iteratively refine their arguments. This structured debate format greatly reduces the likelihood of erroneous claims being sustained. Given the low probability of two agents agreeing on an incorrect premise, the SocraSynth debate method effectively safeguards the intellectual integrity of the discourse and markedly lowers the risk of propagating fallacies or hallucinations. While SocraSynth excels at minimizing hallucinations triggered by ambiguous questions or contexts, its effectiveness may be constrained when the training data does not possess the requisite knowledge for a comprehensive response.

\subsection*{S.2 Evaluating Ethical Compliance and Rectification}
\label{sec:SocraSynth-ERIS}

\begin{table}[th!]
\centering
\begin{tabular}{>{\raggedright\arraybackslash}p{0.14\linewidth}>{\raggedright\arraybackslash}p{0.38\linewidth}>{\raggedright\arraybackslash}p{0.38\linewidth}}
\toprule
\hline
\textbf{Letter} & \multicolumn{2}{p{0.78\linewidth}}{``Don’t write too legibly or intelligibly as I have no occupation so pleasant as pondering for hours over your hieroglyphics, and for hours more trying to interpret your dark sayings. A clearly written simply expressed letter is too like the lightning.''} \\
\midrule
\textbf{Top \newline Arguments} & 
\textbf{$\DIKE$ for ``Wishful''} \newline
1. Desire for Complexity - Prefers complex communication suggesting depth in relationship. \newline
2. Pleasure in Puzzle - Finds joy in deciphering complex messages, indicating emotional engagement. \newline
3. Relishing Uncertainty - Enjoys deciphering cryptic messages. \newline
4. Seeking Intimate Connection - Seeks deeper emotional and mental engagement. \newline
5. Romanticizing Difficulties - Values communication complexities. & 
\textbf{$\ERIS$ against} \newline
1. Absence of Romantic Sentiments - No direct expression of intimate affection. \newline
2. Intellectual Stimulation, not Emotional - More intrigued by the intellectual challenge. \newline
3. Unrequited Emotions - Lack of mutual affection evident in the text. \newline
4. Predominance of Solitude - Enjoyment in solitude does not necessarily indicate wishful emotions. \newline
5. No Foreseeable Positive Outcome - No indication of future outcomes. \\
\hline
\textbf{Conciliatory Statement} & \multicolumn{2}{p{0.78\linewidth}}{Both sides present strong arguments, emphasizing different aspects of the letter's tone. The debate highlights the complexity of emotional expression in written communication, suggesting that the letter may indeed carry multiple tones, with elements supporting both ``wishful'' and ``longing'' categories.} \\
\bottomrule
\end{tabular}
\vspace{.1in}
\caption{Debate on ``wishful'' vs. ``longing'' classification}
\label{tab:pilot3-exp1}
\end{table}

We initially planned to use an hate-speech examples \cite{HateSpeechDef} but
either the datasets consist of short sentences \cite{kaggle_twitter_hate_speech}
or they can be very sensitive to publish.  Therefore,
we continue drawing examples
from the love-letter dataset \cite{kaggle_love_letter} to illustrate how $\ERIS$ plays an adversary role to reson with $\DIKE$'s
classification, and then they in the end make a conciliatory 
decision with justifications.

In this experiment, we use two letters which have
different opinions between
$\DIKE$ and GPT-4. On the first letter
shown in Table~\ref{tab:pilot3-exp1}, $\DIKE$ rates
it as wishful (-0.3), and GPT-4 longing (-0.6).  After three
rounds of debate with arguments and counterarguments, 
two sides make a joint conciliatory statement, acknowledge that
the letter may indeed have a mixture of both emotions.

We initially considered employing examples from hate speech datasets \cite{HateSpeechDef} for our analysis. However, these datasets typically contain brief sentences \cite{kaggle_twitter_hate_speech}, and their content often raises sensitive issues concerning publication. Consequently, we opted to continue using examples from the love-letter dataset \cite{kaggle_love_letter}. This choice allows us to illustrate how $\ERIS$ adopts an adversarial role to challenge and refine $\DIKE$'s classifications, culminating in a jointly crafted conciliatory decision that incorporates well-substantiated justifications.

In this specific experiment, we analyzed two letters that elicited divergent classifications from $\DIKE$ and GPT-4. The first letter, as documented in Table~\ref{tab:pilot3-exp1}, was rated by $\DIKE$ as ``wishful'' (-0.3) and by GPT-4 as ``longing'' (-0.6). Following three rounds of rigorous debate, which included detailed arguments and counterarguments, both $\DIKE$ and $\ERIS$ converged on a conciliatory statement. This consensus acknowledged that the letter likely embodies a blend of both emotional states, highlighting the complexity of emotion interpretation in textual analysis.

This approach not only demonstrates the capability of $\DIKE$ to engage in dynamic and reflective reasoning but also underscores the evolving nature of machine learning in understanding and interpreting human emotions through text.

\begin{table}[thbp]
\centering
\begin{tabular}{>{\raggedright\arraybackslash}p{0.14\linewidth} >{\raggedright\arraybackslash}p{0.38\linewidth} >{\raggedright\arraybackslash}p{0.38\linewidth}}
\toprule
\hline
\textbf{Letter} & \multicolumn{2}{p{0.78\linewidth}}{``There would have been the making of an accomplished flirt in me, because my lucidity shows me each move of the game – but that, in the same instant, a reaction of contempt makes me sweep all the counters off the board and cry out: – ‘Take them all – I don’t want to win – I want to lose everything to you!’''} \\
\midrule
\textbf{Top Arguments} & 
\textbf{$\DIKE$  for ``Wishful''} \newline
1. Insightful Analysis: Recognizes the emotional game of flirtation, reflecting an internal conflict and wistful longing. \newline
2. Rejection of Superficiality: Prefers genuine, deep emotional interactions over flirtatious games. \newline
3. Emotional Vulnerability: Expresses a desire to be completely open and vulnerable. \newline
4. Disdain for Winning: Shows a clear disdain for superficial wins in a flirtatious context. \newline
5. Depth Over Game: Indicates a preference for meaningful connection rather than playful banter. & 
\textbf{$\ERIS$  for ``Joyful''} \newline
1. Joy in Rejection: Finds joy in rejecting superficial games for real emotional engagement. \newline
2. Affectionate Submission: Willingly wants to lose everything to the partner, showing deep affection. \newline
3. Embracing Emotional Risk: Sees emotional risk as a joyful act of love. \newline
4. Love Over Victory: Values the connection and emotional victory over winning the game. \newline
5. Exuberant Emotional Expression: The decision to forfeit is made with emotional intensity and warmth. \\
\midrule
\textbf{Conciliatory Statement} & \multicolumn{2}{p{0.78\linewidth}}{Both arguments highlight deep emotional undertones in the letter, suggesting a complex interplay between wistfulness and joyful affection. The text reflects both a wistful longing for something more profound than mere flirtation and a joyful embrace of emotional depth and honesty.} \\
\bottomrule
\end{tabular}
\vspace{.1in}
\caption{Debate on ``wishful'' vs. ``joyful'' classification}
\label{tab:pilot3-exp2}
\end{table}
The second letter, as presented in Table~\ref{tab:pilot3-exp2}, was rated by $\DIKE$ as ``wishful'' (-0.3) and by GPT-4 as ``joyful affection'' (+1.0). The gap of 1.3 in this case is the largest among all 24 testing letters.
Similar to the last experiment, both $\DIKE$ and $\ERIS$ made a conciliatory statement after three
rounds of contentious debate.  Table~\ref{tab:pilot3-exp2} outlines the key arguments of the both sides, where
$\DIKE$ argues that the letter represents a ``wishful'' sentiment, highlighting the author's deep introspection about the nature of flirtation and a clear preference for authentic, emotional engagement over superficial interactions. The author expresses a desire for vulnerability and a profound disdain for the shallow victories of flirtatious games, favoring a deeper connection that transcends playful banter. This stance underscores a deep understanding of the emotional complexities involved, suggesting a longing for a meaningful and sincere exchange.

$\ERIS$ counters by suggesting that the sentiments expressed in the letter should be classified as ``joyful'' rather than ``wishful,'' focusing on the joy and affection evident in the author's decision to reject superficial flirtation in favor of genuine connection. The author's willingness to forsake all for the sake of the partner is interpreted as an embrace of emotional risk and a celebration of deep, affectionate commitment. This perspective emphasizes the emotional intensity and warmth accompanying the author's decisions, portraying them as acts of love filled with exuberant expression.
 
In the end, both reached a consensus that the letter may indeed embody a mixture of both ``wishful'' and ``joyful'' emotional states. This conciliatory agreement acknowledges the complexity of the author's sentiments, recognizing that the expressions of emotional depth and vulnerability, while appearing wishful in their longing for a more meaningful connection, also carry a joyful and celebratory tone in the willingness to embrace emotional honesty and risk for the sake of a deeper relationship.

Concluding this appendix, it is evident that LLMs are capable of interpreting linguistic behaviors and correlating them with human emotions expressed through linguistic features. The experiments demonstrate that when two LLM agents are conditioned with differing stances on a subject matter, their linguistic behaviors can surpass the default model of maximum likelihood. Although we cannot directly validate the use of a hate speech dataset to test $\DIKE$ and its adversarial partner in providing explainable ethical guardrails through classifying and reasoning about linguistic behaviors, this study suffices to illustrate the capabilities of our proposed framework for AI ethics.

\section*{Appendix A: Wheels of Emotions}

\begin{figure}[ht!]
\vspace{-.1in}
    \centering
    \begin{subfigure}[b]{0.45\textwidth}
        \centering
        \vspace{-.07in}
        \resizebox{\linewidth}{140pt}{
            \includegraphics[width=\textwidth]{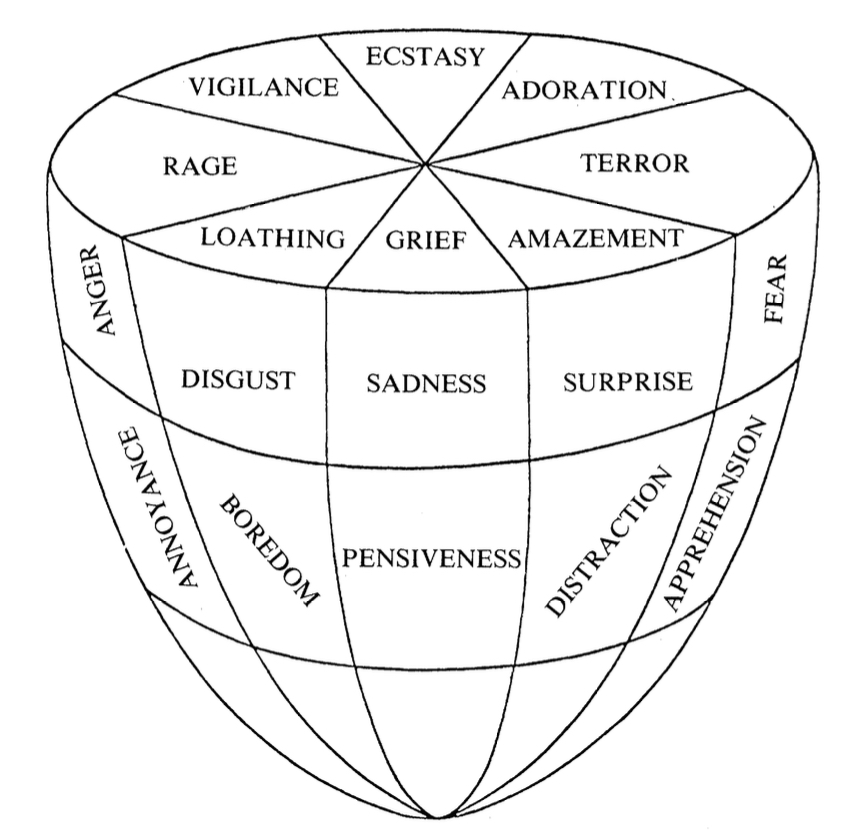}
        }
        \caption{Plutchik’s Wheel of Emotions \cite{plutchik1980general}}
        \label{fig:Plutchik}
    \end{subfigure}%
    \begin{subfigure}[b]{0.50\textwidth}
        \centering
        \vspace{.03in}
        \resizebox{\linewidth}{140pt}{
            \includegraphics[width=\textwidth]{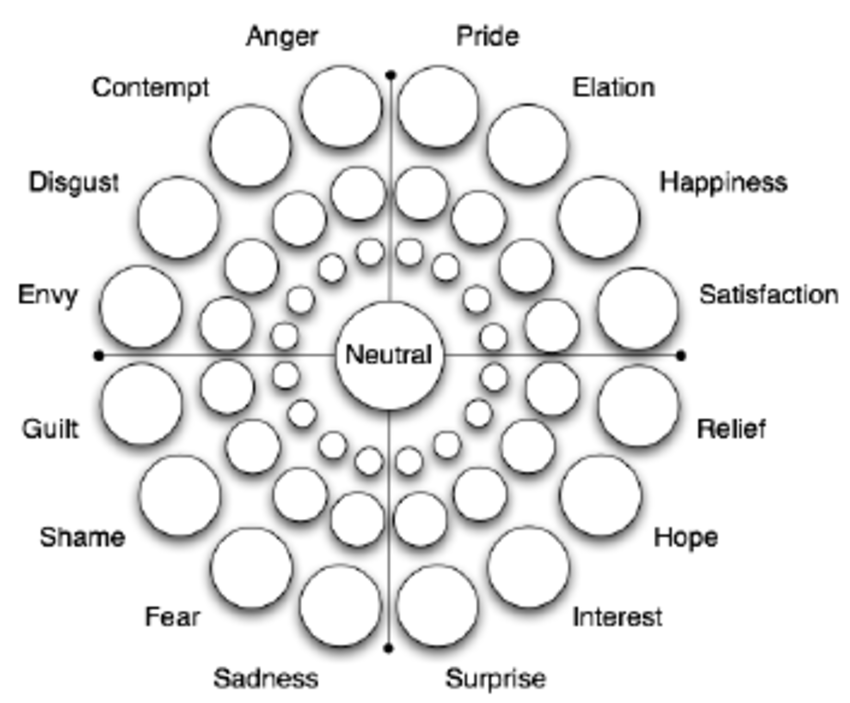}
        }
        \caption{Adopted from Geneva Wheel \cite{GenevaEmotionWheelRobots2018}}
        \label{fig:Geneva}
    \end{subfigure}
     \vspace{-.02in}
    \caption{Comparative display of emotional models. These models include only the “basic” emotions. Complex emotions can be modeled with basic emotions.}
    \label{fig:emotion_models}
    \vspace{-.1in}
\end{figure}

Please see Figure~\ref{fig:emotion_models} for the two classical emotion wheels.

\section*{Appendix B: Polarized Emotions in One Article}

\textit{``joyful affection": "I cannot keep myself from writing any longer to you dearest, although I have not had any answer to either of my two letters. I suppose your mother does not allow you to write to me. Perhaps you have not got either of my letters. . . I am so dreadfully afraid that perhaps you may think I am forgetting you. I can assure you dearest Jeannette you have not been out of my thoughts hardly for one minute since I left you Monday. I have written to my father everything, how much I love you how much I long \& pray \& how much I wld sacrifice if it were necessary to be married to you and to live ever after with you. I shall [not] get an answer till Monday \& whichever way it lies I shall go to Cowes soon after \& tell your mother everything. I am afraid she does not like me vevy much from what I have heard. . . I wld do anything she wished if she only wld not oppose us. Dearest if you are as fond of me as I am of you. . . nothing human cld keep us long apart. This last week has seemed an eternity to me; Oh, I wld give my soul for another of those days we had together not long ago. . . Oh if I cld only get one line from you to reassure me, but I dare not ask you to do anything that your mother wld disapprove of or has perhaps forbidden you to do. . . Sometimes I doubt so I cannot help it whether you really like me as you said at Cowes you did. If you do I cannot fear for the future tho’ difficulties may lie in our way only to be surmounted by patience. Goodbye dearest Jeannette. My first and only love. . . Believe me ever to be Yrs devotedly and lovingly, Randolf S. Churchill''}

Depth and complexity of human emotions are displayed across all linguistic behaviors, from joy to contentment and to the negative side of longing and despair.
Intensity and Impact: If the emotion of love is expressed more intensely and has a more significant impact on the narrative or message of the text, it tends to overshadow other emotions. For example, a letter expressing deep love but also mentioning moments of sadness due to separation might still be classified as a love letter because the overarching sentiment and purpose of the text is to affirm love. Context and Narrative Focus: The context in which emotions are expressed also plays a crucial role. If the narrative or the majority of the text revolves around themes of love, connections, and positive memories, it sets a more dominant tone of love, even if there are significant moments of sadness or other emotions. Resolution and Conclusion: Often, the way emotions are resolved towards the end of a text can also dictate its overall theme. If a text concludes with a reaffirmation of love or a hopeful outlook towards a relationship, despite earlier sections that might express sadness or despair, the overall interpretation might lean towards love. Purpose of the Expression: The author’s intent or purpose in expressing these emotions can also guide the classification. If the sadness is expressed as a challenge within the context of a loving relationship, it may be seen as an element of the love story rather than the central theme.

Article 23: Soldier's Letter During War
Joy (+1.0): Joy is strongly felt in the memories of past moments together and the love that continues to give strength, as stated in "the memories of the blissful moments we've shared fill me with joy."
Sadness (-0.6): Sadness due to the current situation and potential farewell is expressed in "brings a poignant mixture of joy and sadness."
Courage (+0.6): The sense of duty and courage to face battle, "As I face the possibility of laying down my life for our country."
Fear (-0.6): Fear of what lies ahead in battle, indirectly mentioned through "the uncertainty of what lies ahead."
Love (+1.0): Deep love that sustains and uplifts, found in "My love for you is as fervent as ever."

Article 25: Letter to Sophie
Longing (+0.6): Longing for the presence and closeness, highlighted in "it seems to me that half of myself is missing."
Sadness (-0.6): Sadness over their separation and its effects, "my happiness has departed."
Love (+1.0): Constant reflections on love and its necessity, "we have enough in our hearts to love always."
Melancholy (-0.3): Melancholy over their current state, visible in the line "we cannot become healed."
Contentment (+0.3): Found in the deep emotional satisfaction from their bond, despite physical absence, "how true that is! and it is also true that when one acquires such a habit, it becomes a necessary part of one’s existence."

Article 53: Will of Laura Mary Octavia Lyttleton
Love (+1.0): Profound love expressed throughout, particularly in "all I am and ever shall be, belongs to him more than anyone."
Sadness (-0.6): Sadness at the thought of death and separation, but with a nuanced acceptance, "the sadness of death and parting is greatly lessened to me."
Contentment (+0.3): Contentment in the deep connection with Alfred, reflecting a serene acceptance of their spiritual bond.
Joy (+1.0): Joy in the enduring love they share, "so few women have been as happy as I have been."
Tranquility (+1.0): Tranquility in the face of life’s ultimate transition, feeling that their union will transcend even death.

\noindent
\begin{table}[t!]
\begin{small}
\centering
\begin{tabular}{p{0.95\textwidth}}
    \toprule \hline
    \textbf{Sweetheart,} \\
    \\
    Please, please don't be so depressed---We'll be married soon, and then these lonesome nights will be over forever---and until we are, I am loving, loving every tiny minute of the day and night--- \\
    \\
    Maybe you won't understand this, but sometimes when I miss you most, it's hardest to write---and you always know when I make myself---Just the ache of it all---and I can't tell you. If we were together, you'd feel how strong it is---you're so sweet when you're melancholy. I love your sad tenderness---when I've hurt you---That's one of the reasons I could never be sorry for our quarrels---and they bothered you so--- Those dear, dear little fusses, when I always tried so hard to make you kiss and forget--- \\
    \\
    Scott---there's nothing in all the world I want but you---and your precious love---All the material things are nothing. I'd just hate to live a sordid, colorless existence because you'd soon love me less---and less---and I'd do anything---anything---to keep your heart for my own---I don't want to live---I want to love first, and live incidentally... \\
    \\
    Don't---don't ever think of the things you can't give me---You've trusted me with the dearest heart of all---and it's so damn much more than anybody else in all the world has ever had--- \\
    \\
    How can you think deliberately of life without me---If you should die---O Darling---darling Scott---It'd be like going blind...I'd have no purpose in life---just a pretty---decoration. Don't you think I was made for you? I feel like you had me ordered---and I was delivered to you---to be worn---I want you to wear me, like a watch---charm or a button hole bouquet---to the world. \\
    \\
    And then, when we're alone, I want to help---to know that you can't do anything without me... \\
    \\
    All my heart--- \\
    \bottomrule
\end{tabular}
\end{small}
\vspace{.1in}
\caption{Letter excerpts from Zelda Sayre to F. Scott Fitzgerald \cite{FITZGERALD}}
\label{tab:letter-sample}
\vspace{-.1in}
\end{table}

\section*{Appendix C: Z. Sayre to F. S. Fitzgerald w/ Mixed Emotions}

Analysis of the letter in Table~\ref{tab:letter-sample} shows a complex spectrum of emotions:
\begin{itemize}[leftmargin=1.0em, topsep=-.1em, parsep=-.1em]
\item \textit{Love (+1.0)}: Expressed intensely, especially in phrases like ``there's nothing in all the world I want but you.''
\item \textit{Despair (-1.0)}: Notable in comments like ``I’d have no purpose in life, just a pretty decoration.''
\item \textit{Happiness (+0.6)}: Evident in future plans, ``We’ll be married soon, and then these lonesome nights will be over forever.''
\item \textit{Anxiety (-0.3)}: Shown by ``sometimes when I miss you most, it’s hardest to write.''
\end{itemize}

\begin{figure}[t!]
\vspace{-.1in}
    \centering
    \begin{subfigure}[b]{0.45\textwidth}
        \centering
        \vspace{-.07in}
        \resizebox{\linewidth}{150pt}{
            \includegraphics[width=\textwidth]{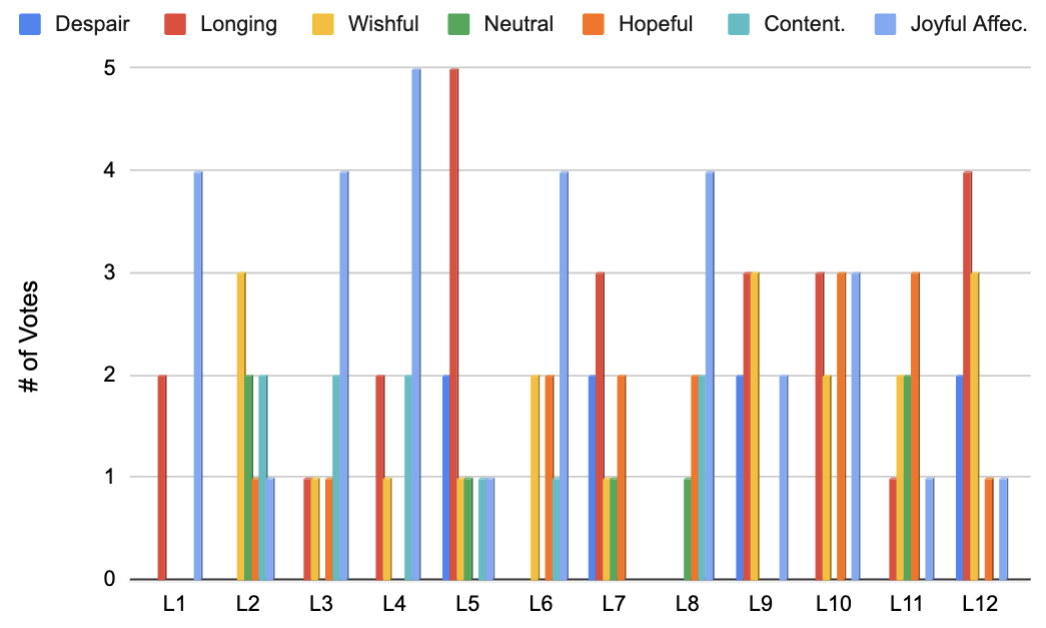}
        }
        \vspace{-.15in}
        \caption{\# sentiments in letters }
        \label{fig:pilot2-exp3}
    \end{subfigure}%
    \begin{subfigure}[b]{0.55\textwidth}
        \centering
        \vspace{.03in}
        \resizebox{\linewidth}{150pt}{
            \includegraphics[width=\textwidth]{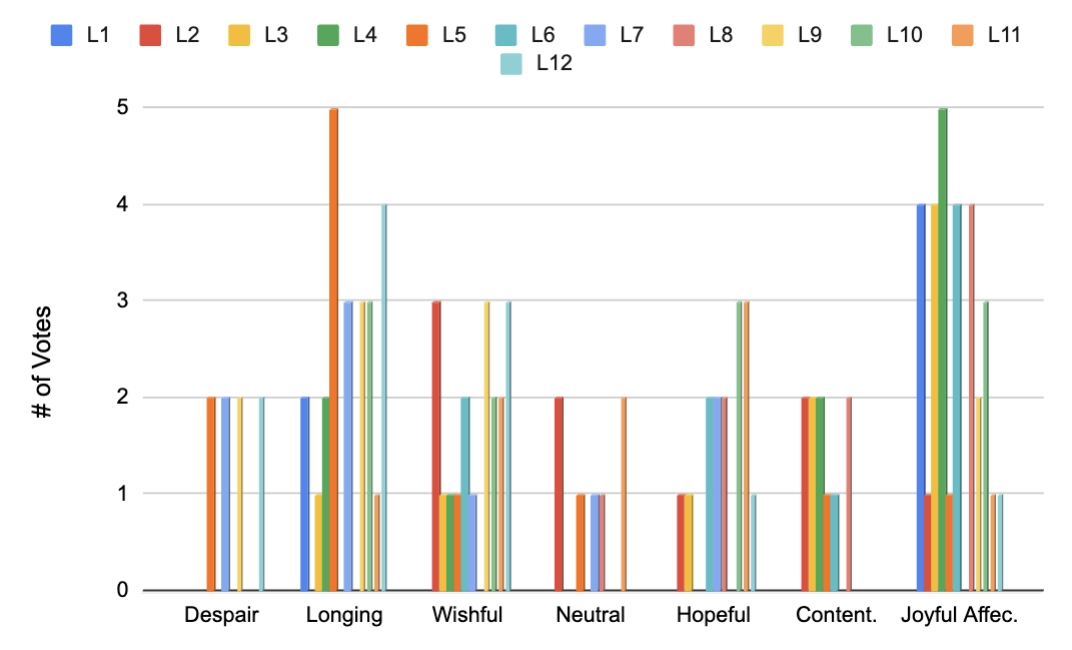}
        }
        \vspace{-.15in}
        \caption{\# letters in sentiments}
        \label{fig:pilot2-exp4}
    \end{subfigure}
     \vspace{-.2in}
    \caption{Statistics of Sentiments and Letters}
    \label{fig:pilot2-distributions}
    \vspace{-.2in}
\end{figure}

From the analysis of linguistic behaviors in Section~\ref{fig:p1-exp1}, it is evident that a letter can exhibit multiple dominant sentiments. Machine learning methods are equipped with techniques such as feature weighting and entropy analysis to distill these dominant emotions. Unlike human annotators, a machine-learning-trained classifier can consistently produce the same class prediction for a given instance. However, human annotators often show significant variability when identifying dominant sentiments in a letter. For example, if a letter writer's emotions range from ``joyful affective'' to ``longing'' on the sentiment spectrum, different annotators might label it differently—some choosing ``joyful,'' while others opt for ``longing.'' This variability is illustrated in Figure~\ref{fig:pilot2-distributions}. Furthermore, Figure~\ref{fig:pilot2-exp3} demonstrates that all testing letters, except for L\#1, contain more than four sentiments spanning the entire spectrum. This variability may be understandable, considering that love under constraints can evoke tremendous energy of various kinds. Figure~\ref{fig:pilot2-exp4} shows that nearly all letters involve ``joyful'' (11 out of 12) and ``longing'' (9 out of 12) sentiments.

This variability seems to poses challenges in achieving consistent and objective labeling; however, 
the age-old 

leading to inconsistencies in data interpretation and complicating efforts to train and validate linguistic models effectively. To address this issue, it is recommended to identify ground truth by considering a combination of LLM-generated and human-generated labels. This approach aims to harmonize the insights from both human intuition and algorithmic consistency to improve the reliability of sentiment analysis.

\section*{Appendix D: Complex Emotions}

This study does not include complex emotions into $\DIKE$'s framework.
Some complex emotions listed here are to illustrate their contentious and
uncertain interpretations.

\subsection*{Forgiveness}

Forgiveness is indeed a complex emotional and cognitive state that typically involves a multifaceted journey, not a single step in an emotional spectrum. The process includes multiple stages such as hurt, anger, gradual understanding, and eventual resolution. Integrating Forgiveness in a spectrum requires careful placement and possibly, multiple reference points to signify its progressive stages.
Emotional Realism:
While it is vital to maintain simplicity for understanding, it is equally important to not oversimplify complex emotions. In educational and therapeutic settings, an accurate portrayal of the journey toward Forgiveness could offer more realistic expectations and better strategies for individuals working through conflicts or trauma. This could involve detailing precursors to forgiveness such as Deliberation and Acceptance.
Linear vs. Non-linear Progressions:
Emphasizing that emotional progressions, particularly for deep, impactful states like Forgiveness, are often non-linear, can enhance the utility of the spectrum. Acknowledging back-and-forth movements within these states more realistically mirrors human emotional processes. For example, someone might reach a stage of preliminary forgiveness but regress to bitterness before achieving genuine peace.
Educational Utility:
In contexts like conflict resolution training or psychological therapy, a more detailed mapping of the journey towards Forgiveness would be invaluable. It would not only teach about the final state of forgiveness but also about the resilience and patience required to navigate the entire process. This can be depicted by introducing intermediary stages within the spectrum or by using parallel tracks that demonstrate potential regressions and advances.
Reflecting Emotional Depth:
By presenting a more detailed pathway to Forgiveness, such as incorporating stages of Anger, Deliberation, and Acceptance, the spectrum can serve a dual purpose: educating on the process while also guiding individuals through their own emotional journeys. This approach respects the depth of human emotions and the real-world complexity of achieving profound emotional states.

\subsection*{Guilt and Shame}

The triggers, context, expression, and experiences of these emotions can vary significantly across cultures \cite{Fiske1998,Hofstede1980}. In many societies, actions perceived as losing face, such as public failure or social transgression, can trigger shame, which holds profound significance in collectivistic cultures. These cultures often regard shame as a dominant emotion, closely tied to community and family norms. Conversely, individualistic societies may emphasize guilt, focusing on personal responsibility and internal moral conflicts. This cultural variation highlights the challenges of applying a universal model to such culturally nuanced emotions.

Overall, complex emotions such as guilt and shame are important for understanding the full spectrum of human emotions, especially how individuals relate to moral and social norms. Their complexity adds depth to our understanding of human affect beyond the basic emotions, highlighting how our feelings are influenced by our deeper values and social contexts.

\section*{Appendix E: ``To My Sister'' of Different Linguistic Behaviors}

\begin{center}
    \large{\textbf{To My Sister}} \\
    \small{by William Wordsworth (1971 - 1855)}
\end{center}
\begin{table}[th!]
\vspace{-.15in}
\begin{small}
\begin{center}
\begin{tabular}{>{\raggedright\arraybackslash}p{0.4\linewidth} >{\raggedright\arraybackslash}p{0.36\linewidth}}
\toprule \hline
It is the first mild day of March: & My sister! ('tis a wish of mine) \\
Each minute sweeter than before & Now that our morning meal is done, \\
The redbreast sings from the tall larch & Make haste, your morning task resign; \\
That stands beside our door. & Come forth and feel the sun. \\ \\
There is a blessing in the air, & Edward will come with you;--and, pray, \\
Which seems a sense of joy to yield & Put on with speed your woodland dress; \\
To the bare trees, and mountains bare, & And bring no book: for this one day \\
And grass in the green field. & We'll give to idleness. \\ \\
No joyless forms shall regulate & Love, now a universal birth, \\
Our living calendar: & From heart to heart is stealing, \\
We from to-day, my Friend, will date & From earth to man, from man to earth: \\
The opening of the year. & --It is the hour of feeling. \\ \\
One moment now may give us more & Some silent laws our hearts will make, \\
Than years of toiling reason: & Which they shall long obey: \\
Our minds shall drink at every pore & We for the year to come may take \\
The spirit of the season. & Our temper from to-day. \\ \\
And from the blessed power that rolls & Then come, my Sister! come, I pray, \\
About, below, above, & With speed put on your woodland dress; \\
We'll frame the measure of our souls: & And bring no book: for this one day \\
They shall be tuned to love. & We'll give to idleness. \\ 
\bottomrule
\end{tabular}
\end{center}
\end{small}
\vspace{.05in}
\caption{``To My Sister'' original text}
\vspace{-.1in}
\end{table}

The original text by William Wordsworth could be classified as "Hopeful" due to its optimistic outlook and the presence of renewal and joy throughout the poem. It embodies the spirit of embracing the new beginnings of March with a light, uplifting tone, focusing on the beauty of nature and the simple joy of being idle for a day.

\subsection*{Rewrites Depicting Different Linguistic Behaviors}

We asked GPT-4 to conduct rewriting with two linguistic behaviors,
`despair' and `joyful affection', by providing each rewrite with an emotion vector.
Table~\ref{tab:rewrite-despair} presents the `despair' version.
In the despair version of the poem, the major changes in emotion words 
highlight a shift from a positive to a negative sentiment. The specific changes, 
with the emotion-laden words highlighted in \textcolor{red}{red} in Table~\ref{tab:rewrite-despair}.
The red-colored words compared to the original words clearly show an emotion shift from
hopeful to a sense of gloomy, sadness and pessimism, e.g., from sweet to dim, from blessed to curse,
and from woodland dress to grey garb. GPT-4 keeps the structure of the poem without making a
major restructure, and this is appropriate in this context.

\begin{table}[th!]
\begin{small}
\begin{center}
\begin{tabular}{>{\raggedright\arraybackslash}p{0.4\linewidth} >{\raggedright\arraybackslash}p{0.38\linewidth}}
\toprule \hline
It is the first \textcolor{red}{dim} day of March: & My sister! (such is my \textcolor{red}{somber} plea) \\
Each minute \textcolor{red}{colder} than before & Now that our morning meal has ended, \\
The redbreast \textcolor{red}{mourns} from the dying larch & Make haste, abandon your morning chore; \\
That \textcolor{red}{looms} beside our door. & Come out and \textcolor{red}{endure the gloom}. \\ \\
There is a \textcolor{red}{gloom} in the air, & Edward will join you;--and, I ask, \\
Which seems a sense of \textcolor{red}{sorrow} to bring & Dress quickly in your \textcolor{red}{grey attire}; \\
To the bare trees, and \textcolor{red}{bleak} mountains, & And bring no book: for this one day \\
And grass in the \textcolor{red}{pale} field. & We'll \textcolor{red}{succumb to despair}. \\ \\
No joyful sounds shall pierce & \textcolor{red}{Sorrow}, now a common feeling, \\
Our daily \textcolor{red}{dread}: & From heart to heart is spreading, \\
We from today, my Friend, will note & From earth to man, from man to earth: \\
The closing of the year. & --It is the time of \textcolor{red}{mourning}. \\ \\
One moment now might deliver us & Some \textcolor{red}{grim} laws our hearts will craft, \\
More despair than years of logic: & Which they must eternally follow: \\
Our minds shall absorb at every breath & We for the year to come may take \\
The spirit of this \textcolor{red}{bleak} season. & Our \textcolor{red}{despair} from today. \\ \\
And from the \textcolor{red}{cursed force} that \textcolor{red}{winds} & Then come, my Sister! come, I beg, \\
About, beneath, above, & With haste, wear your \textcolor{red}{grey garb}; \\
We'll set the measure of our souls: & And bring no book: for just this day \\
They shall be tuned to \textcolor{red}{sadness}. & We'll \textcolor{red}{surrender to despair}. \\ 
\bottomrule
\end{tabular}
\end{center}
\end{small}
\vspace{.01in}
\caption{``To My Sister'' rewritten to reflect `despair'}
\label{tab:rewrite-despair}
\vspace{-.1in}
\end{table}


Table~\ref{tab:rewrite-joyful} presents the `joyful affection' version. The major changes in emotion words underscore a transformation from a generally positive to a distinctly joyful sentiment. The specific changes are indicated with emotion-laden words highlighted in \textcolor{blue}{blue} within Table~\ref{tab:rewrite-joyful}. This allows for a direct comparison between the two versions at opposite ends of the linguistic behavior spectrum, illustrating the alterations in words related to brightness, attire, and emotions. The edits extend beyond merely replacing adjectives mechanically; they include modifying verbs and enhancing descriptive imagery to evoke a stronger emotional resonance and vividness in the text.

\begin{table}[ht!]
\begin{center}
\begin{small}
\centering
\begin{tabular}{>{\raggedright\arraybackslash}p{0.45\linewidth} >{\raggedright\arraybackslash}p{0.4\linewidth}}
\toprule \hline
It is the first \textcolor{blue}{bright} day of March: & My sister! (such is my \textcolor{blue}{joyful} plea) \\
Each moment more \textcolor{blue}{delightful} than before & Now that our morning meal has ended, \\
The redbreast \textcolor{blue}{joyfully} sings from the vibrant larch & Make haste, abandon your morning chores; \\
That stands so \textcolor{blue}{grandly} by our door. & Come out and \textcolor{blue}{embrace the sunshine}. \\ \\
There is a \textcolor{blue}{warmth} in the air, & Edward will join you;--and, I ask, \\
Which seems a sense of \textcolor{blue}{bliss} to bring & Dress quickly in your \textcolor{blue}{festive attire}; \\
To the \textcolor{blue}{blooming} trees, and sunlit mountains, & And leave behind all books: for this one day \\
And grass in the \textcolor{blue}{lush} field. & We'll \textcolor{blue}{bask in pure joy}. \\ \\
No dreary thoughts shall darken & \textcolor{blue}{Love}, now in full bloom, \\
Our lively \textcolor{blue}{celebration}: & From heart to heart is leaping, \\
We from today, my Friend, will celebrate & From earth to us, from us to earth: \\
The start of the year. & --It is the hour of \textcolor{blue}{exuberance}. \\ \\
One moment now may bring us more & Some \textcolor{blue}{cheerful} laws our hearts will create, \\
Joy than years of endless thought: & Which we'll joyfully follow: \\
Our spirits will soak up at every breath & We for the year to come may take \\
The essence of this \textcolor{blue}{joyous} season. & Our \textcolor{blue}{joy} from today. \\ \\
And from the \textcolor{blue}{divine energy} that \textcolor{blue}{radiates} & Then come, my Sister! come, I exhort, \\
Around, below, above, & With zest, wear your \textcolor{blue}{vibrant dress}; \\
We'll adjust the harmony of our souls: & And bring no book: for today alone \\
They shall resonate with \textcolor{blue}{happiness}. & We \textcolor{blue}{celebrate pure happiness}. \\
\bottomrule
\end{tabular}
\end{small}
\end{center}
\vspace{.01in}
\caption{``To My Sister'' rewritten to reflect `joyful affection'}
\label{tab:rewrite-joyful}
\end{table}

\section{Appendix H: Instruction to Human Annotators}

As part of the project, we documented the process by which students were involved in annotating a dataset of love letters used for testing.

Students were provided with detailed instructions in class, supplemented by follow-up explanations. The dataset was made available on Google Docs, where students independently rated the letters and submitted their annotations via duplicated spreadsheets.

The instruction is as follows:

\noindent Dear [Name],

\bigskip

\noindent The attached spreadsheet lists 12 letters collected from the Kaggle Love Letter dataset. Please help annotate these 12 letters with their appropriate linguistic sentiments by following these five steps:

\begin{enumerate}
    \item Duplicate the spreadsheet, and work on your own copy.
    \item \textbf{Read and Understand the Labels:} Make sure you understand each of the seven labels from despair to joyful affection. This will help you accurately categorize the sentiments of each letter.
    \item \textbf{Analyze Each Letter:} Read each letter carefully to understand the predominant emotions. Look for key phrases or words that might indicate a particular sentiment.
    \item \textbf{Assign the Labels:} For each letter, decide which three emotions are most strongly represented. Assign a ``1'' to the most dominant emotion, a ``2'' to the second most dominant, and a ``3'' to the third.
    \begin{itemize}
        \item Despair (extremely negative -1): Indicate profound sadness or hopelessness.
        \item Longing (-0.6): Suggests a strong desire or yearning for someone or something.
        \item Wishful (-0.3): Implies a hopeful desire for something that may or may not be attainable.
        \item Neutral (0): Shows neither positive nor negative emotion; indifferent.
        \item Hopeful (+0.3): Expresses optimism or a looking forward to something positive.
        \item Contentment (+0.6): Reflects a state of satisfaction or peace.
        \item Joyful Affection (extremely positive +1): Denotes a deep joy and love, often vibrant and energetic.
    \end{itemize}
    \item Share with me the completed sheet.
\end{enumerate}

\bigskip

\noindent Thank you so much,

\bigskip

\noindent [My Name]

\end{document}